\pdfoutput=1

\documentclass[11pt]{article}

\usepackage[]{acl}

\usepackage{enumitem}
\usepackage{microtype}
\usepackage{makecell}
\usepackage{inconsolata}
\usepackage{colortbl}

\usepackage{times}
\usepackage{latexsym}
\usepackage{amsmath}

\usepackage{times}
\usepackage{latexsym}
\usepackage[T1]{fontenc}

\usepackage[T1]{fontenc}

\usepackage[utf8]{inputenc}

\usepackage{microtype}

\usepackage{inconsolata}

\usepackage{graphicx}
\usepackage{amsfonts}
\usepackage{amsmath}

\usepackage{booktabs}
\usepackage{multirow}
\usepackage{amsmath}
\usepackage{microtype}
\usepackage{graphicx}
\usepackage{amsfonts}
\usepackage{tabularx}
\usepackage{color}
\usepackage{comment}
\usepackage{amsmath,amsfonts,amssymb}
\usepackage{arydshln}
\usepackage{tablefootnote}
\usepackage{xcolor}
\usepackage{subcaption}
\usepackage{float}

\usepackage{bbding}
\usepackage{pifont}
\usepackage{wasysym}
\usepackage{utfsym}
\usepackage{fontawesome}

\usepackage{amsthm}

\title{Finding the Sweet Spot: Preference Data Construction for Scaling Preference Optimization}

\author{
Yao Xiao$^{1,4,}$\thanks{This work was partially done during the internship of YX at MiroMind.}
Hai Ye$^{2,4}$
Linyao Chen$^{3}$\\
\textbf{
Hwee Tou Ng$^{2}$
Lidong Bing$^{4}$
Xiaoli Li$^{5}$
Roy Ka-Wei Lee$^{1}$}
\\
$^1$Singapore University of Technology and Design\\
$^2$National University of Singapore\\
$^3$The University of Tokyo
$^4$MiroMind\\
$^5$Institute for Infocomm Research, A*Star, Singapore\\
\\
}

\begin{document}
\maketitle

\begin{abstract}

Iterative data generation and model retraining are widely used to align large language models (LLMs).
It typically involves a policy model to generate on-policy responses and a reward model to guide training data selection. 
Direct Preference Optimization (DPO) further enhances this process by constructing preference pairs of chosen and rejected responses. 
In this work, we aim to \emph{scale up} the number of on-policy samples via repeated random sampling to improve alignment performance. 
Conventional practice selects the sample with the highest reward as chosen and the lowest as rejected for DPO. 
However, our experiments reveal that this strategy leads to a \emph{decline} in performance as the sample size increases. 
To address this, we investigate preference data construction through the lens of the underlying normal distribution of sample rewards. 
We categorize the reward space into seven representative points and systematically explore all 21 ($C_7^2$) pairwise combinations. 
Through evaluations on four models using AlpacaEval 2, we find that selecting the rejected response at reward position $\mu - 2\sigma$, rather than the minimum reward, is crucial for optimal performance. 
We finally introduce a scalable preference data construction strategy that consistently enhances model performance as the number of samples scales up.\footnote{Our code and data can be found at \url{https://github.com/XYaoooo/DPO_Pair}.}


\end{abstract}

\section{Introduction}
\label{sec:intro}

Large language models (LLMs) have significantly advanced natural language processing, demonstrating remarkable capabilities across various tasks~\cite{NEURIPS2020_1457c0d6, wei2022finetuned, bubeck2023sparksartificialgeneralintelligence}.  
However, these models still generate unintended outputs due to their unsupervised nature~\cite{bai2022traininghelpfulharmlessassistant, 10.1145/3531146.3533088}.  
To mitigate these issues, recent efforts have focused on improving LLM alignment with human preferences~\cite{NEURIPS2022_b1efde53, rafailov2023direct, pmlr-v238-gheshlaghi-azar24a, pmlr-v235-ethayarajh24a, meng2024simpo}. 

Reinforcement learning from human feedback (RLHF) has become a widely adopted framework to align LLMs with human preferences.  
RLHF involves first training a reward model, which then provides feedback signals to optimize a policy model through reinforcement learning, typically using Proximal Policy Optimization (PPO)~\cite{schulman2017proximalpolicyoptimizationalgorithms, ahmadian-etal-2024-back}.  
However, this approach is complex and often unstable as it requires substantial memory and computational resources to accommodate three separate models simultaneously~\cite{xu2024is, ivison2024unpacking}. 
To address these challenges, \citet{rafailov2023direct} introduced Direct Preference Optimization (DPO), which bypasses the need for reward models.
Subsequent research has focused on improving preference optimization efficiency~\cite{liu2024statistical, pmlr-v238-gheshlaghi-azar24a, pmlr-v235-ethayarajh24a, han2024fpogeneralizingpreferenceoptimization}.

\begin{figure*}[!t]
\centering
\includegraphics[width=0.98\linewidth, clip=true, trim = 20mm 50mm 22mm 88mm]{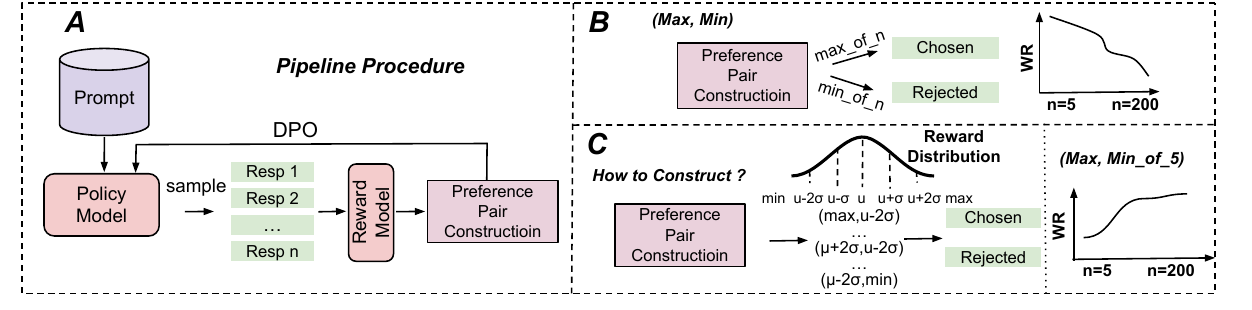}
\caption{We orderly show (A) the procedure of preference data construction; (B) the conventional preference data construction strategy; (C) our exploration and proposed method. }
\label{pipe}
\vspace{-1em}
\end{figure*}

Currently, with the increasing availability of powerful reward models~\cite{pmlr-v202-gao23h, wang-etal-2024-interpretable, liu2024skyworkrewardbagtricksreward}, an effective pipeline (Figure~\ref{pipe}) to further enhance the alignment capabilities of LLMs without human annotations has gained popularity.  
To start with, $n$ on-policy responses~\cite{tajwar2024preferencefinetuningllmsleverage, guo2024directlanguagemodelalignment, song2024the, 
  tang2025rlfinetuningllmsonoffpolicy} are first sampled from LLMs and subsequently scored by a reward model.
The response with the highest reward is selected as the chosen response, while the one with the lowest reward is selected as the rejected response to construct a preference dataset.
The constructed preference dataset can be used to train the policy model through DPO in return. 
In practice, five samples per prompt can achieve significant performance gains~\cite{meng2024simpo}.

\begin{figure*}[!t]
\centering
\subfloat[Llama]{%
\begin{minipage}[t]{0.5\linewidth} 
    \centering
    \includegraphics[width=0.45\linewidth]{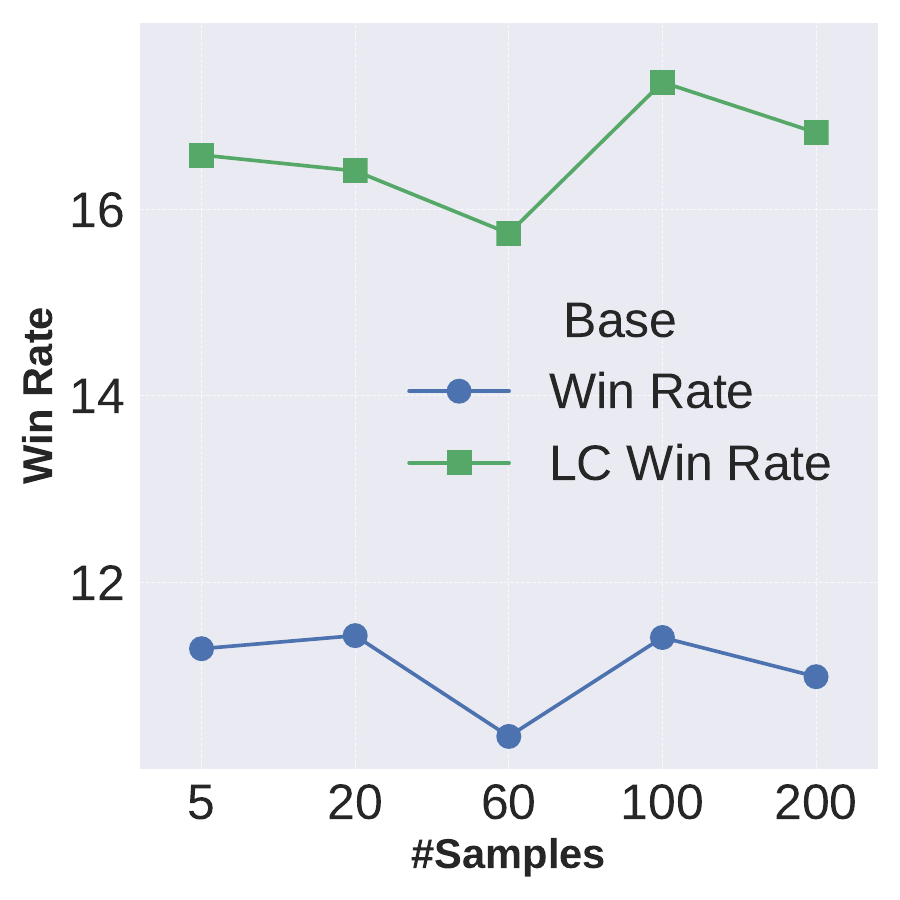}
    \hfill
    \includegraphics[width=0.45\linewidth]{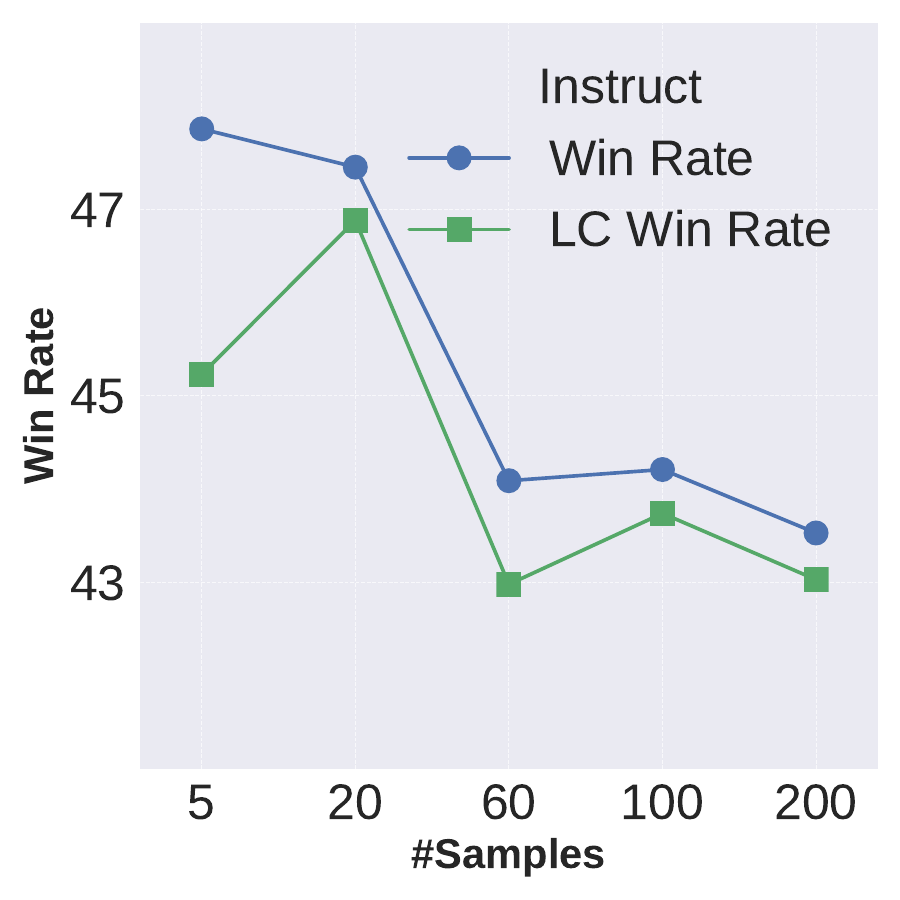}
\end{minipage}%
}
\hfill
\subfloat[Mistral]{%
\begin{minipage}[t]{0.5\linewidth} 
    \centering
    \includegraphics[width=0.45\linewidth]{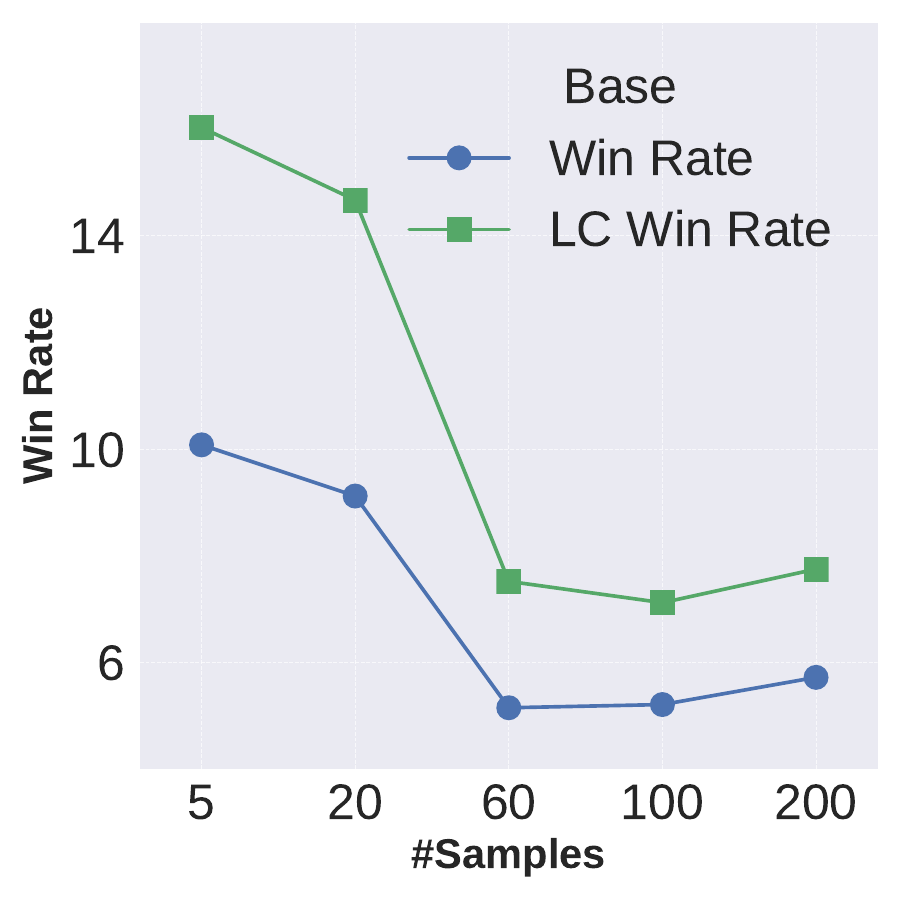}
    \hfill
    \includegraphics[width=0.45\linewidth]{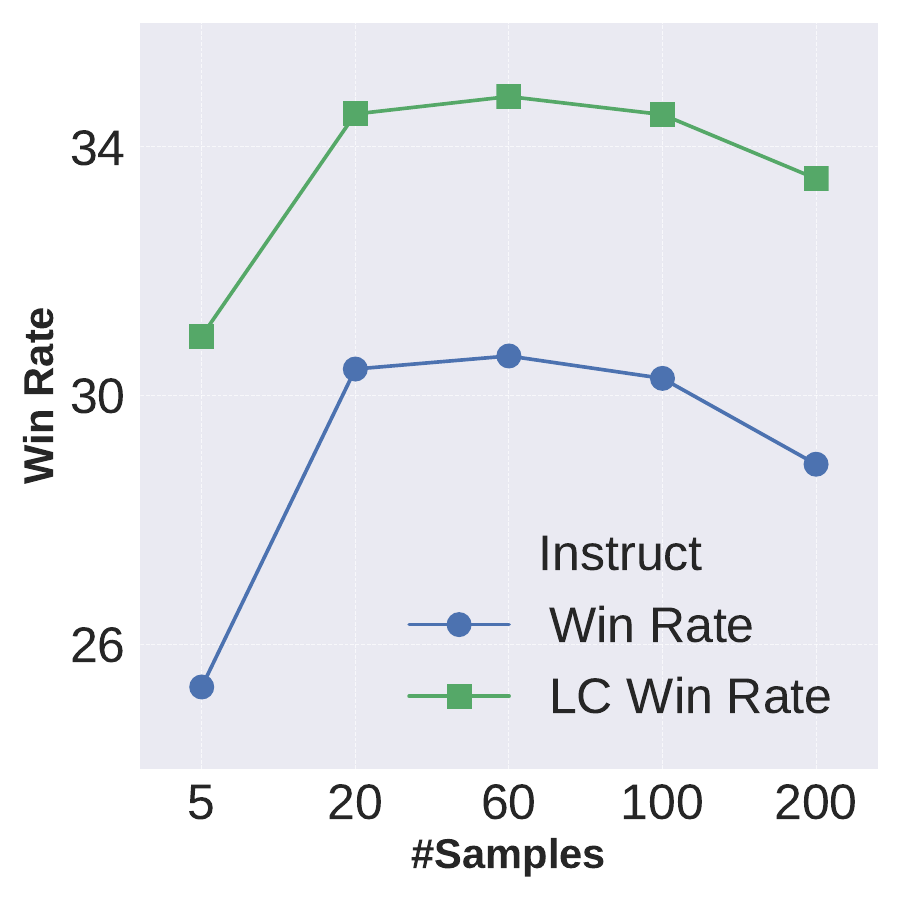}
\end{minipage}%
}
\caption{Alpaca evaluation results. The conventional approach which selects the response with the highest reward as the chosen response and the response with the lowest reward as the rejected response for DPO fails to improve the performance of models when we increase the number of samples. The x-axis represents the number of samples, while the y-axis shows the score (\%).}
\label{dy_wr}
\vspace{-1em}
\end{figure*}

In this paper, we focus on the construction of preference pairs~\cite{kim-etal-2025-systematic} given sufficient sample budgets by scaling up the number of samples per prompt.
We first construct preference pairs following the pipeline described above and train policy models.
However, increasing the number of samples does not lead to any significant performance improvement and can even result in a decline, as shown in Figure~\ref{dy_wr}.
As preference pairs play a pivotal role in DPO training, we then investigate the construction of preference data based on the underlying normal distribution of rewards.
Specifically, we classify the samples per prompt into seven categories and construct 21 preference datasets. 
We systematically explore the performance of trained models through DPO with each preference dataset and observe consistent empirical results across different settings.
Finally, we propose a simple preference data construction strategy that can steadily improve the performance of trained models through DPO if we increase the number of samples.


\paragraph{Contributions.} 
This work makes the following key contributions:
\begin{itemize}
    \item We point out that the conventional preference pair construction strategy fails to improve the performance of models when increasing the number of samples per prompt.
    
    \item We, for the first time, construct preference pairs based on the distribution of rewards and explore their effects on policy models. We find that selecting the rejected response at reward position $\mu - 2\sigma$ is a key factor for optimal results.

    \item We propose a scalable preference pair construction method to improve the performance of models with an increasing number of samples per prompt. Its effectiveness can be further demonstrated by comparing with previous work.
\end{itemize}

    



\section{Background}





\captionsetup[subfigure]{aboveskip=0pt, belowskip=2pt}
\captionsetup[figure]{aboveskip=2pt, belowskip=0pt}

\subsection{Direct Preference Optimization}
Different from conventional RLHF which first compresses human preferences into reward models, direct preference optimization is a RL-free algorithm for training language models to align with human preferences.
DPO is recognized as one of the most widely used methods for preference optimization. 
It reformulates the reward function $r$ into a closed-form expression aligned with the optimal policy:
\begin{equation}
r(x, y) = \beta \log \frac{\pi_\theta(y | x)}{\pi_{\text{ref}}(y | x)} + \beta \log Z(x) \nonumber
\end{equation}

where $\pi_\theta$ denotes the policy model, $\pi_{\text{ref}}$ represents the reference model (usually the supervised fine-tuned model) and $Z(x)$ is the partition function. 
By embedding this reward formulation into the Bradley-Terry (BT) ranking framework, the probability of preference $p(y_w > y_l | x)$ is computed as $\sigma(r(x, y_w) - r(x, y_l))$, where $\sigma$ is the sigmoid function. Therefore, DPO replaces the reliance on a reward model with the policy model, resulting in the following objective:
\begin{align}
& \mathcal{L}_{\text{DPO}}(\pi_\theta; \pi_{\text{ref}}) = \notag \\
& -\mathbb{E}_{(x, y_w, y_l) \sim \mathcal{D}} \Big[ \log \sigma(r(x, y_w) - r(x, y_l)) \Big] \notag
\end{align}

where $r(x, y) = \beta \log \frac{\pi_\theta(y \mid x)}{\pi_{\text{ref}}(y \mid x)}$.

\subsection{Preference Data Construction}
\label{conven_pipe}
Recently, a method for constructing preference pairs without relying on human annotation has been gaining popularity~\cite{dong2023raft, meng2024simpo}.
As shown in Figure~\ref{pipe}, given a language model policy $\pi_{\theta}$,  a reward function $r$ and $k$ prompts $\left\{x_i\right\}_{i=1}^k$, we sample $n$ generations $\left\{y_{ij}\right\}_{j=1}^n$ for the $i$-th prompt from $\pi_{\theta}$. 
The given reward model will be used to score the sampled generations.
The reward of $n$ candidate samples of the $i$-th prompt is $\left\{r_{ij}\right\}_{j=1}^n$.
Afterwards, the completion of the highest reward score $\max_{j=1}^{n} \{r_{ij}\}$
is selected as the chosen response, while the completion of the lowest reward score $\min_{j=1}^{n} \{r_{ij}\}$ is selected as the rejected response to construct a preference pair $(y_w^{(i)}, y_l^{(i)})$ for $x_i$. 
In practice, $n=5$ can achieve significant performance gains~\cite{meng2024simpo}.
In this work, we explore the effects of increasing the number of samples $n$.

\section{Case Study on Conventional Preference Data Construction}

In this part, we follow the preference data strategy described in Section~\ref{conven_pipe} to construct preference data and train policy models.

\subsection{Experimental Setup}
\label{imp_detail}

We follow the setup in \citet{meng2024simpo} to conduct experiments using two setups: \textbf{Base} and \textbf{Instruct}. For Base setting, we first fine-tune Llama-3-8B and Mistral-7B-v0.1 on the UltraChat-200k dataset~\cite{ding2023enhancing} to obtain the SFT model. For Instruct setting, we use Llama-3-8B-Instruct and Mistral-7B-Instruct-v0.2 as the SFT models.  
With these SFT models, we sample responses for prompts from UltraFeedback~\cite{pmlr-v235-cui24f}, constructing preference datasets as described in Section~\ref{conven_pipe}. 
We then use the preference dataset scored with the Absolute-Rating Multi-Objective Reward Model (Armorm)~\cite{wang2024arithmetic} to train the SFT model via DPO~\cite{rafailov2023direct}.  
For sampling, we use a temperature of 0.8 and scale the number of samples from 5 to 200. 
We employ vLLM~\cite{kwon2023efficient} for efficient inference.
Following \citet{meng2024simpo}, we set the evaluation temperature to 0.8 for Llama-3-8B and Llama-3-8B-Instruct, while using a temperature of 0.7 for Mistral-7B-v0.1 and 0.5 for Mistral-7B-Instruct-v0.2. 
Additional training hyperparameters are provided in Appendix~\ref{appendix_hyper}.

We mainly evaluate models on \textbf{AlpacaEval 2}~\cite{alpaca_eval, dubois2024lengthcontrolled}, which is the most widely used benchmark for instruction following.
AlpacaEval 2 consists of 805 questions from multiple domains and tasks, which enables a comprehensive assessment of LLMs.
We report both \textit{win rate} and \textit{length-controlled win rate} results (\textit{LC win rate}) for it.


\subsection{Experimental Results} 
The experimental results are presented in Figure~\ref{dy_wr}. 
As the number of samples increases, the performance of the trained models, measured by both win rate and LC win rate, exhibits fluctuations rather than consistent improvements in the base setting of Llama. 
In the instruct setting of Llama, the degradation is even more pronounced, with performance declining as the sample size increases. 
A similar trend is observed for Mistral-7B-v0.1. 
Mistral-7B-Instruct-v0.2 shows a slight improvement before eventually declining, reinforcing the instability of this conventional preference pair construction method.  

These results indicate that the max-min construction strategy cannot improve alignment of LLMs as the sample size increases.  
This finding highlights a limitation in existing preference construction strategies and motivates the need for alternate approaches to constructing preference pairs, particularly when ample samples are available.

\begin{figure*}[!t]
\centering
\subfloat[Armorm]{%
\begin{minipage}[t]{0.5\linewidth} 
    \centering
    \includegraphics[width=0.49\linewidth]{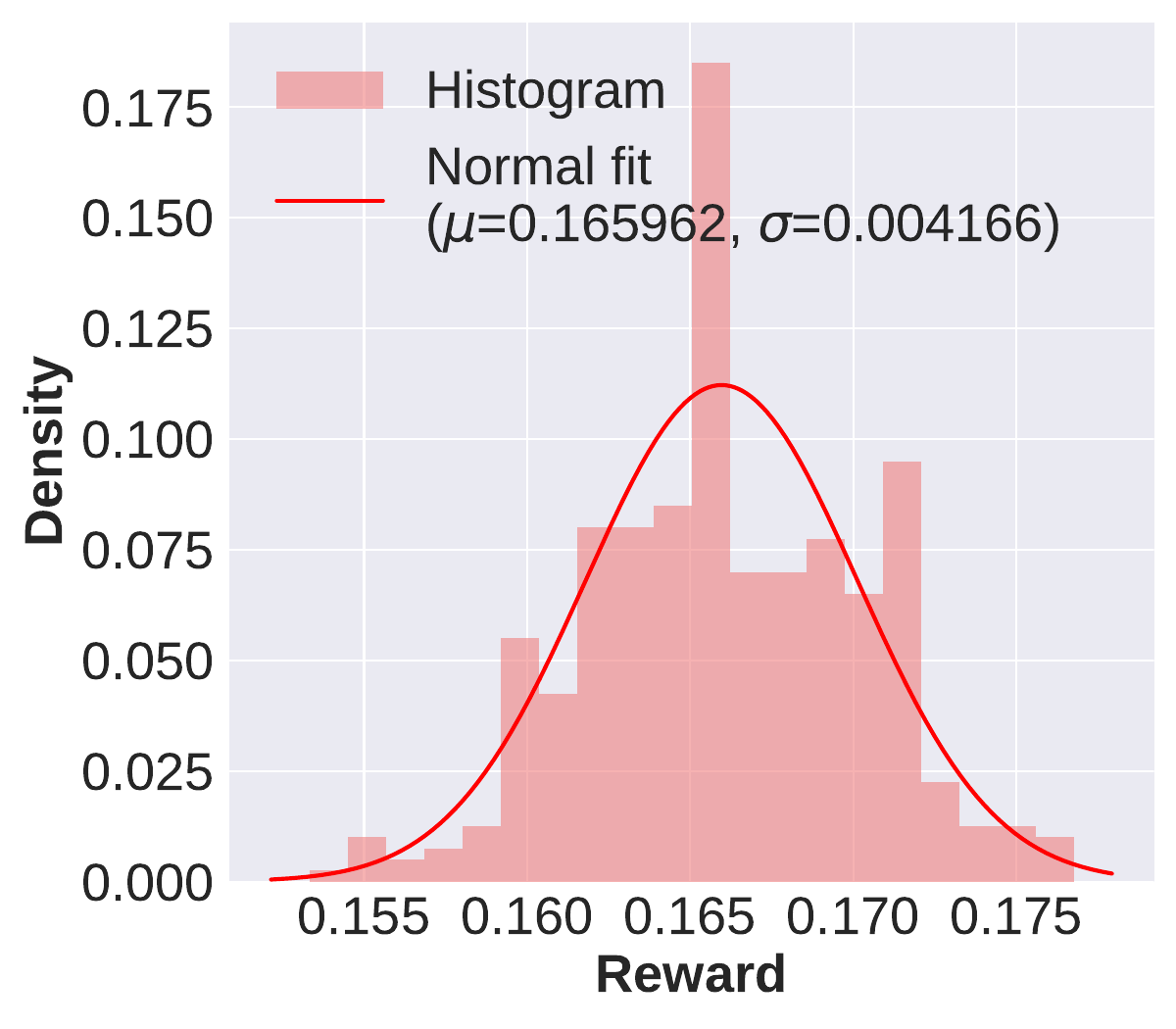}
    \hfill
    \includegraphics[width=0.49\linewidth]{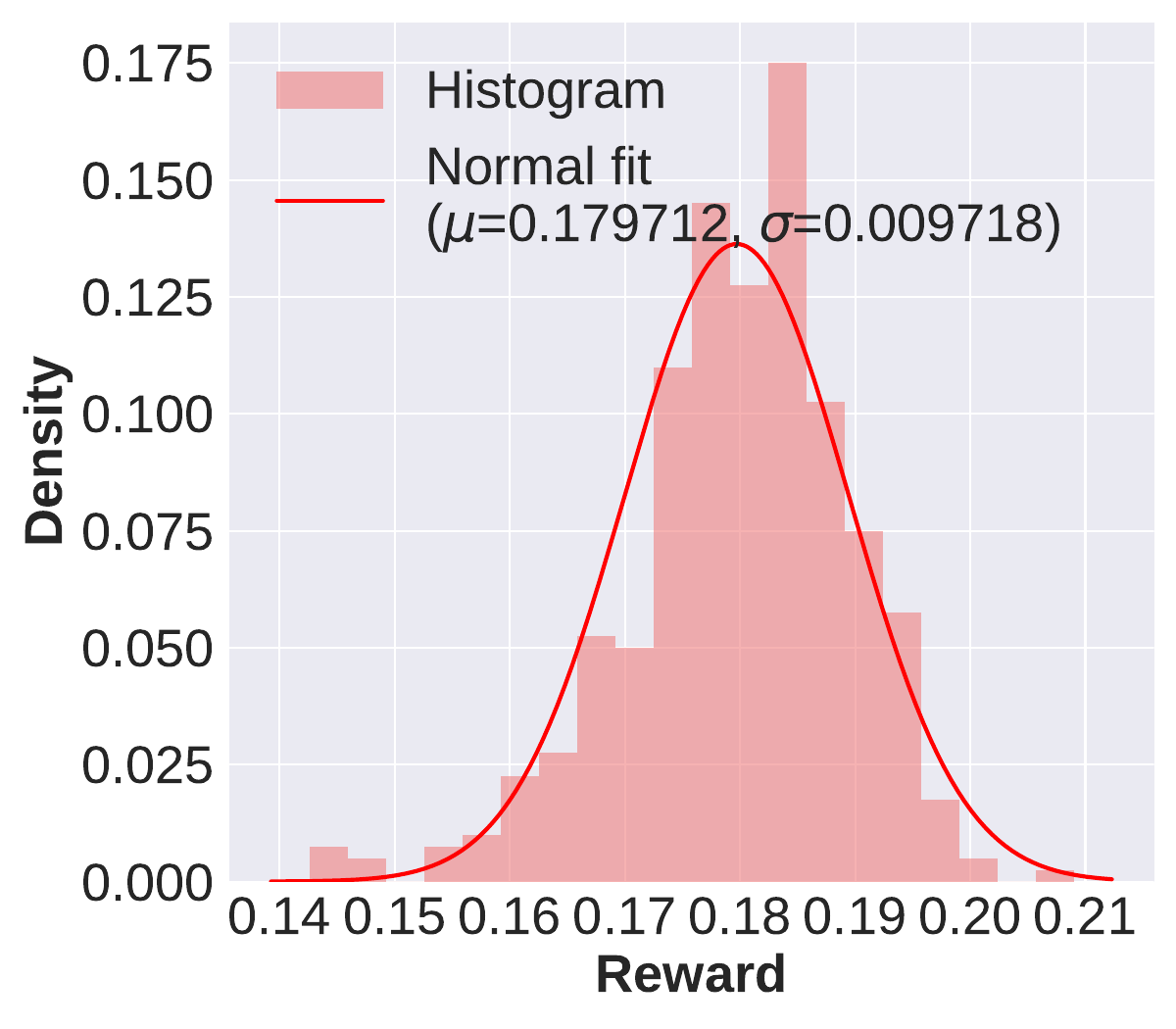}
\end{minipage}%
}
\hfill
\subfloat[Skywork]{%
\begin{minipage}[t]{0.5\linewidth} 
    \centering
    \includegraphics[width=0.49\linewidth]{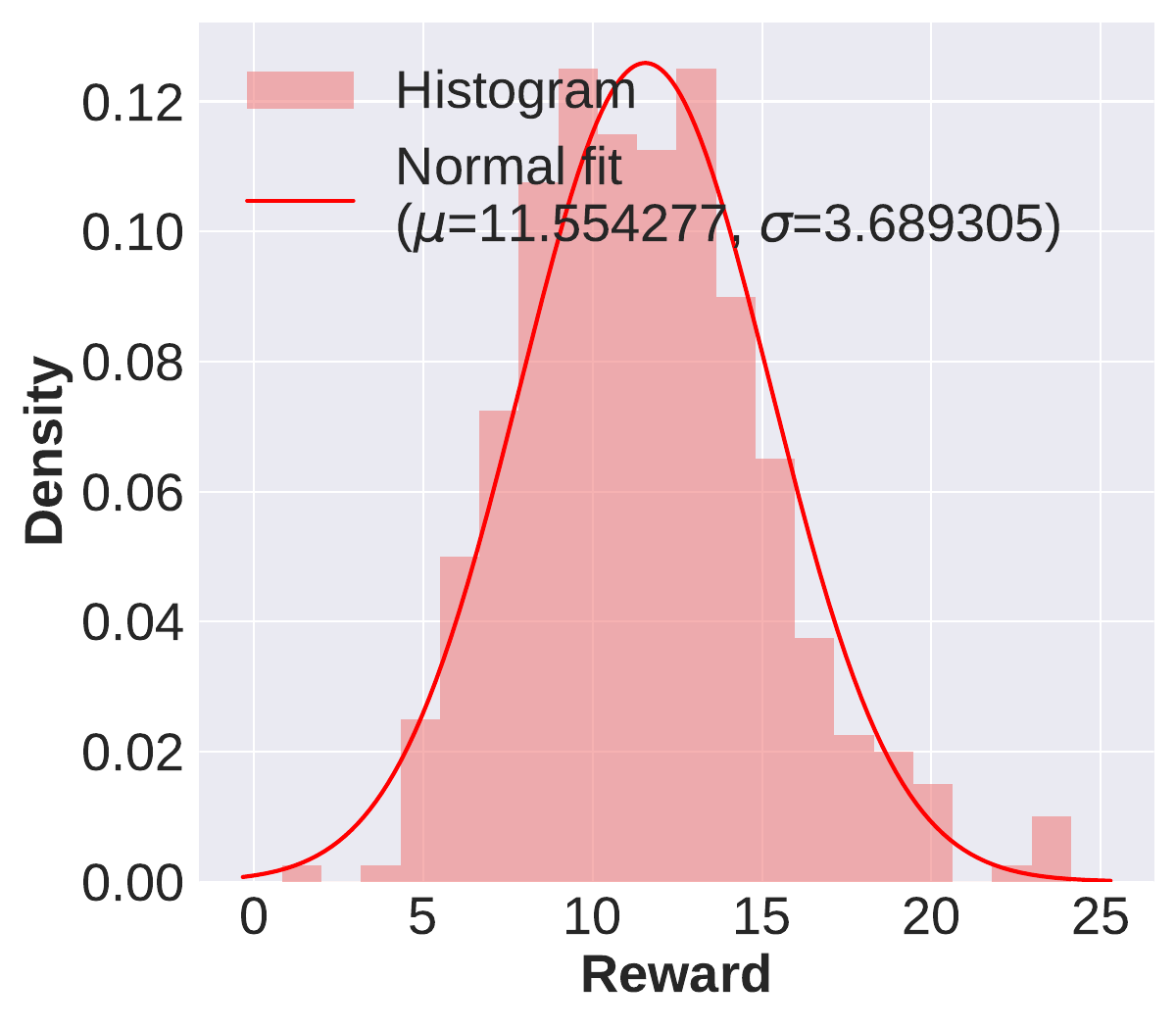}
    \hfill
    \includegraphics[width=0.49\linewidth]{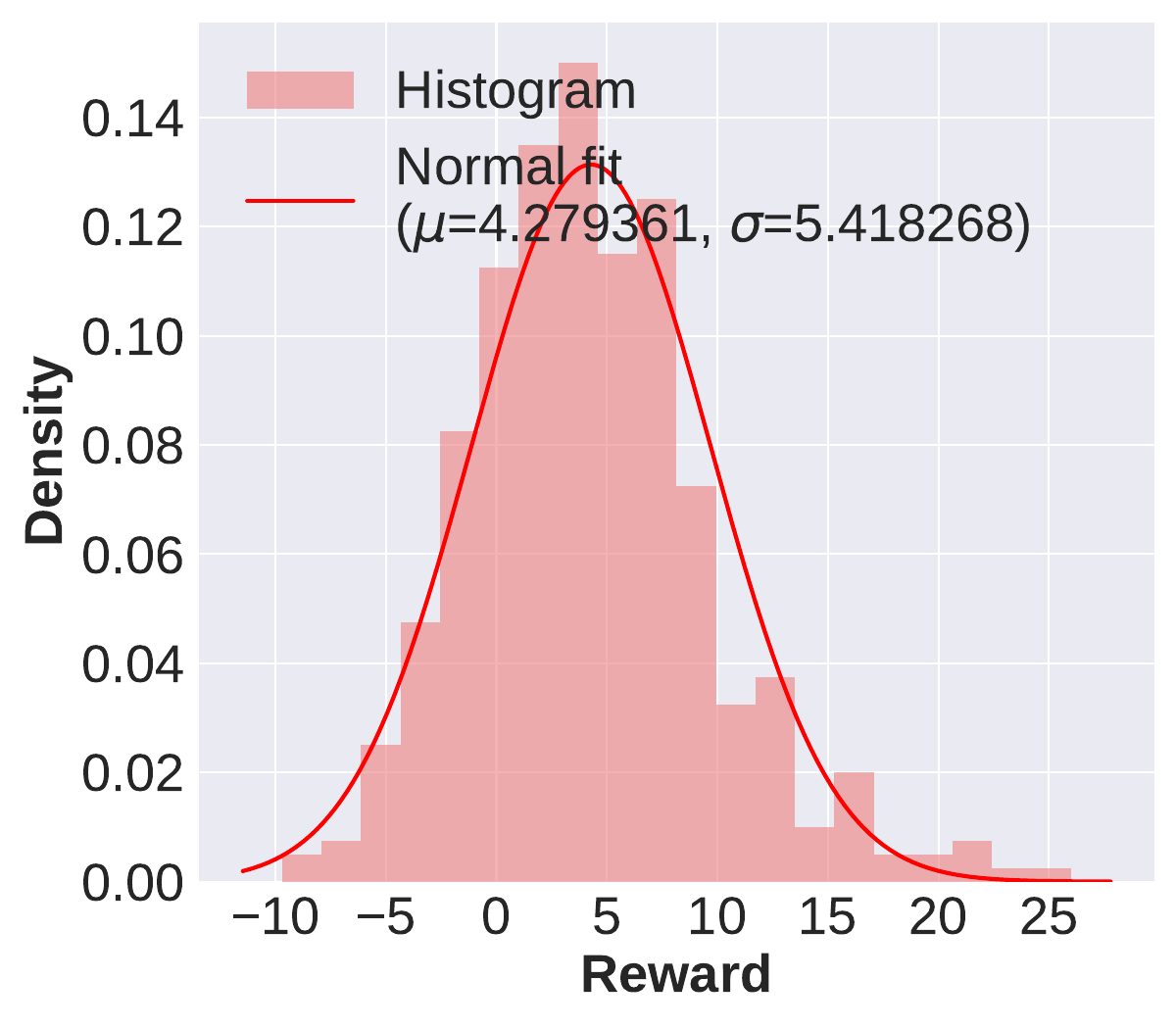}
\end{minipage}%
}
\caption{We plot the reward distribution of 400 completions for 2 random prompts and their approximate Gaussian distribution, calculating with Absolute-Rating Multi-Objective reward model (Armorm)~\cite{wang2024arithmetic} and Skywork reward models (Skywork)~\cite{liu2024skyworkrewardbagtricksreward}.}
\label{normal_dist}
\vspace{-1em}
\end{figure*}

\section{Preference Data Construction via Reward Distribution}
\label{main_method}
In this section, we explore alternative ways to categorize sampled responses based on their reward scores, focusing on a distribution-based approach. 
We first discuss the limitations of ranking-based categorization and introduce a reward distribution-based strategy.
We then describe the preference pair construction process, followed by experimental validation and the key insights derived.


\subsection{Reward Distribution}
In reality, reward scores often exhibit a skewed or clustered distribution, making it challenging to establish meaningful distinctions using fixed ranking intervals.
Instead of dividing samples into equal-sized bins, we define preference categories based on the mean (\(\mu\)) and standard deviation (\(\sigma\)) of the underlying normal distribution, as illustrated in Figure~\ref{normal_dist}. 
This method ensures that preference pairs are drawn from statistically meaningful intervals.

By sampling responses at key points in the distribution, such as \(\mu \pm 2\sigma\), \(\mu \pm \sigma\), and \(\mu\), we can capture variations in reward scores that reflect quality distinctions of responses. 
Another advantage of this approach is that it allows for precise control over the reward margin between chosen and rejected responses. 
By leveraging distribution-aware categorization, we aim to construct preference pairs systematically, enabling a more comprehensive understanding of trained models. 

\begin{figure*}[t]
\centering

    \begin{minipage}[c]{\linewidth}
        \centering
        \includegraphics[width=0.49\linewidth]{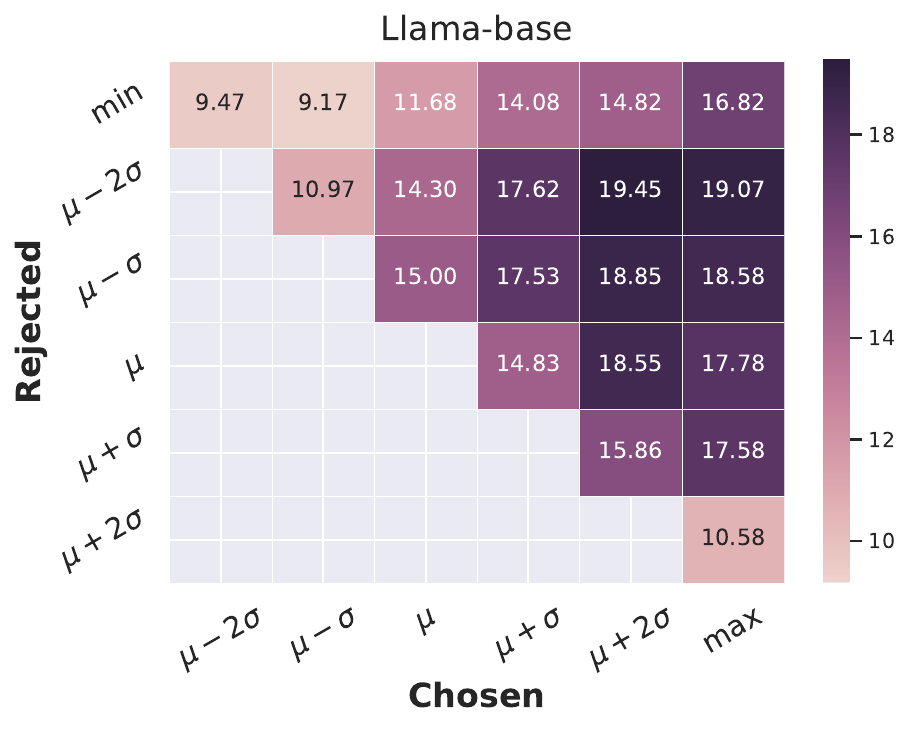}
        \hfill
        \includegraphics[width=0.49\linewidth]{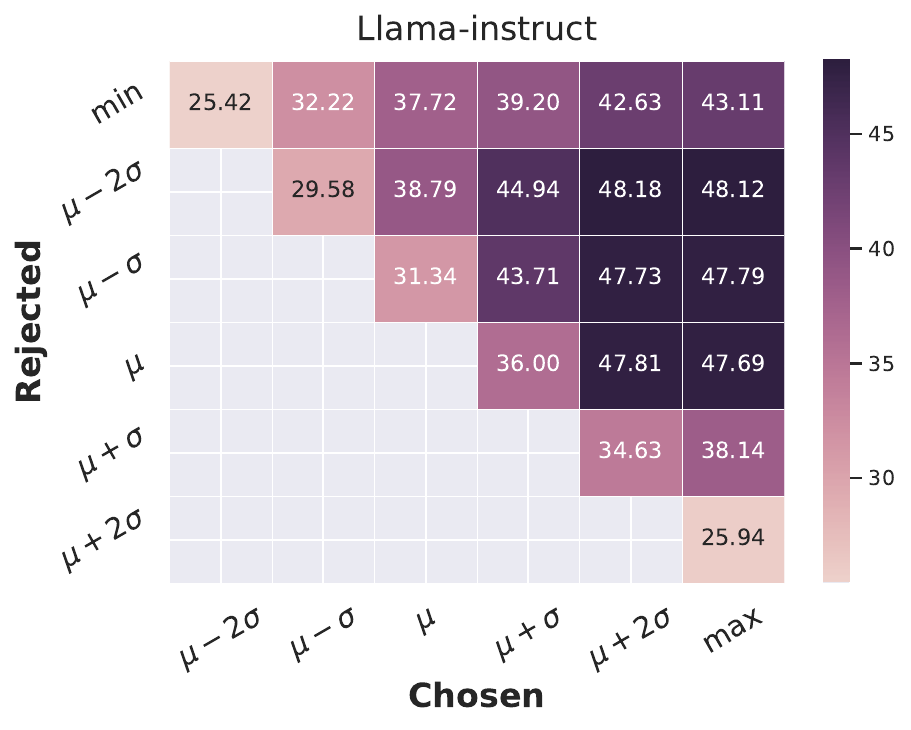}
        \\
        \includegraphics[width=0.49\linewidth]{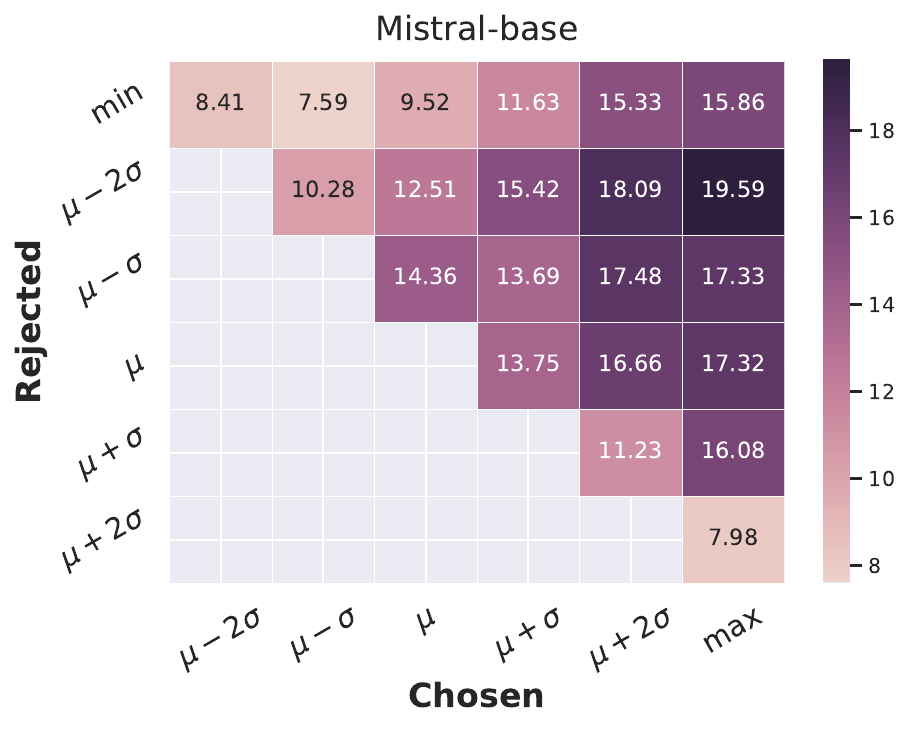}
        \hfill
        \includegraphics[width=0.49\linewidth]{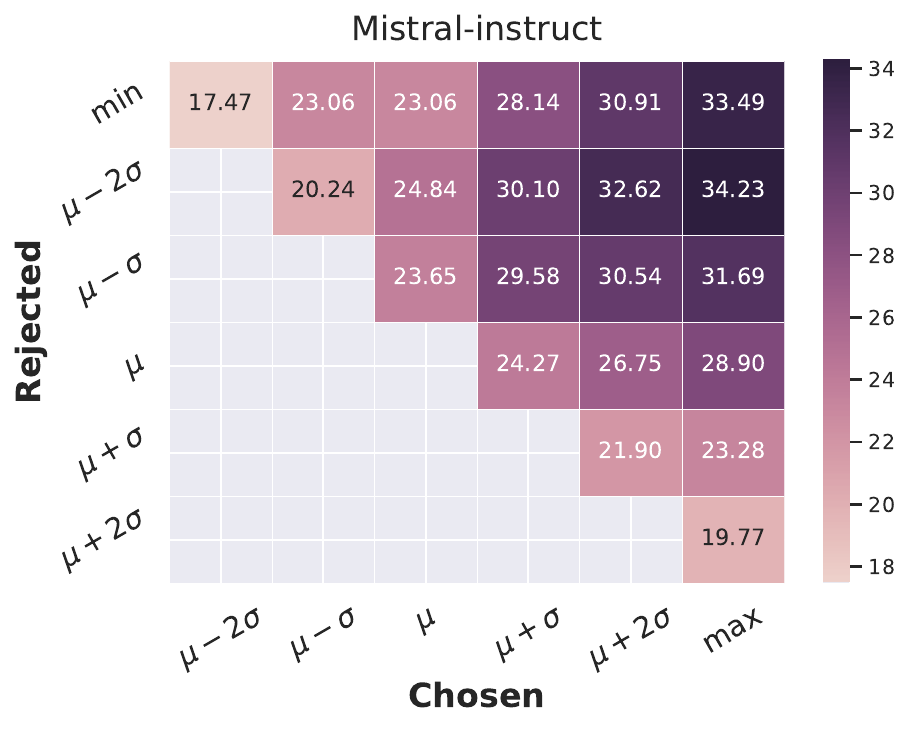}
    \end{minipage}

\caption{Alpaca evaluation results. We report the length-controlled win rate for each preference dataset here. y-axis is the reward point where the rejected response is selected, while x-axis is the reward point where the chosen response is selected.}
\label{main_fig}
\end{figure*}

\subsection{Preference Data Construction}
\label{data_cons}
We propose a structured approach to constructing preference pairs and training policy models through DPO. 
Our method leverages the statistical properties of reward distributions to systematically select responses for preference pair construction.

For each prompt, we first generate \( n \) responses from an SFT model and compute their reward scores using a given reward model. 
Given the reward scores of responses for the \( i \)-th prompt, we approximate the distribution as \( N(\mu_i, \sigma_i^2) \), where \( \mu_i \) and \( \sigma_i \) denote the mean and standard deviation of the rewards, respectively. 
To ensure a representative selection of responses, we extract samples at key points in the reward distribution. 
Specifically, we select responses closest to the values \( \mu_i - 2\sigma_i, \mu_i - \sigma_i, \mu_i, \mu_i + \sigma_i, \mu_i + 2\sigma_i \), along with responses with minimum and maximum reward scores.
This process results in a set of seven different sample points: \( \{ min, \mu \pm 2\sigma, \mu \pm \sigma, \mu, max \} \).
Although \(\mu\) and \(\sigma\) are prompt specific, we drop \(i\) for brevity in the rest of this paper.

The preference pairs are then constructed by considering all possible pairwise combinations of these seven points, following the principle that the chosen response should have a higher reward than the rejected response. 
This results in \( C_7^2 = 21 \) distinct preference pairs per prompt, also 21 preference datasets as a whole. 
We subsequently train 21 different policy models through DPO, each optimized on a unique preference dataset. 
Figure~\ref{pipe} illustrates the overall preference construction process.

\subsection{Experimental Setup}
We largely follow the same experimental and implementation setup described in Section~\ref{imp_detail}. 
We generate 200 samples per prompt and apply the proposed preference data construction strategy. 
For comparison, we also evaluate models trained with conventional preference pair selection, as detailed in Section~\ref{conven_pipe}. 
The results of these baseline models are reported in Appendix~\ref{baseline}.

\subsection{Experimental Results}
We evaluate the performance of 84 policy models trained with the constructed preference datasets, with results presented in Figure~\ref{main_fig}. 
To mitigate biases introduced by response length, we primarily focus on the LC win rate as our evaluation metric~\cite{dubois2024lengthcontrolled}. 
In the following, we summarize our key findings and their implications for preference pair construction in DPO.  

\begin{figure*}[t]
\centering
\includegraphics[width=0.85\linewidth, clip=true, trim = 0mm 0mm 0mm 0mm]{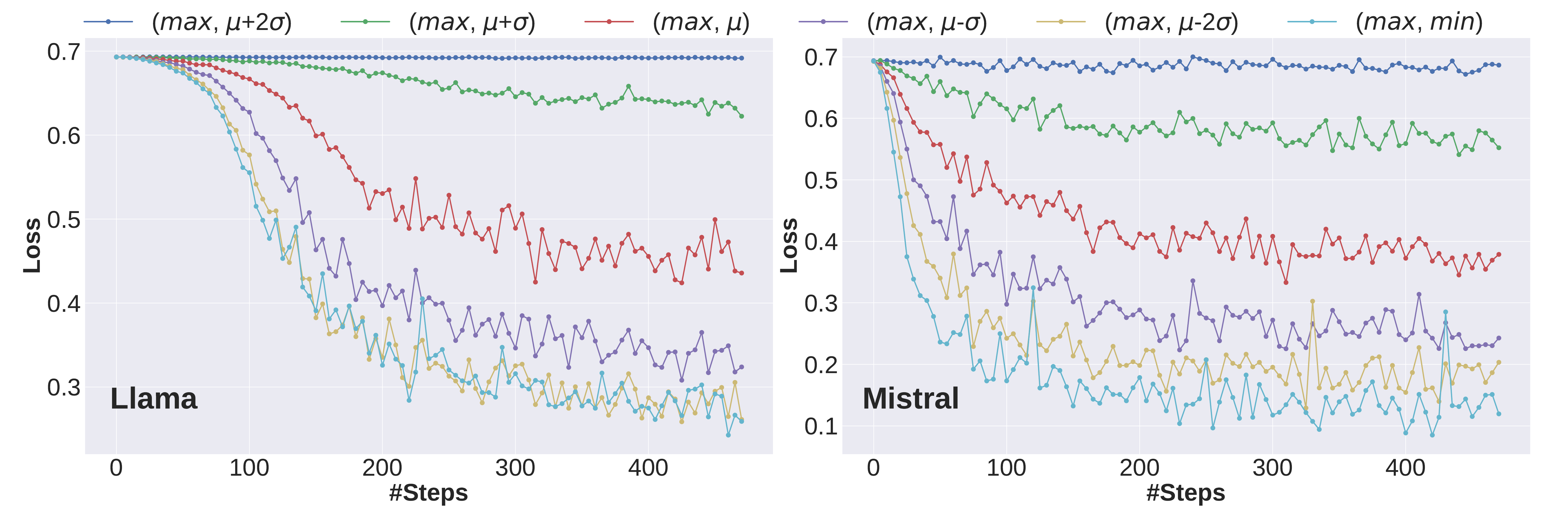}
\caption{We record the training loss for six datasets ($max$, $min$), ($max$, $\mu - 2\sigma$), ($max$, $\mu - \sigma$), ($max$, $\mu$), ($max$, $\mu + \sigma$), and ($max$, $\mu + 2\sigma$) for Llama-3-8B-Instruct and Mistral-7B-Instruct-v0.2 every five steps. x-axis is the step and y-axis is the loss.}
\label{loss}
\vspace{-1em}
\end{figure*}

\paragraph{Impact of Preference Pair Construction on Performance.}
Our results indicate that the chosen response should be selected from \(\{max, \mu+2\sigma\}\).
In addition, the rejected response should be selected at reward position \(\mu-2\sigma\) instead of the minimum reward to produce the optimal performance.  
Among all preference pairs, the pair \((\mu+2\sigma, \mu-2\sigma)\) consistently outperforms others in most cases. 
For example, Llama-3-8B-Instruct trained with this preference pair achieves a length-controlled win rate of 48.18\%, surpassing the conventional preference data construction strategy by about 3 percentage points. 
These findings suggest that preference pairs constructed from well-separated reward intervals improve preference optimization of policy models more effectively than the naive max-min strategy.  

\paragraph{Effect of Reward Margins on Performance.}
A key observation from our experiments is that the performance of trained models improves as the reward of the chosen response increases, provided that the rejected response is appropriately selected. 
When the rejected response is at reward position \(\mu-2\sigma\), the length-controlled win rate increases as the chosen response moves toward higher reward values. 
This trend is witnessed across multiple models and settings, as illustrated in Figure~\ref{main_fig}. 
These results reinforce the importance of ensuring a sufficiently large reward margin between the chosen and rejected responses, which contributes to more effective preference optimization.  

\paragraph{Limitation of Small Reward Margins.}
We further observe that preference pairs with small reward margins perform poorly.  
When the reward of the chosen response is only slightly higher than that of the rejected response, the model struggles to learn meaningful distinctions. 
For example, training Llama-3-8B-Instruct with the pair \((\mu+2\sigma, \mu+\sigma)\) results in a length-controlled win rate of 34.63\%, significantly lower than pairs with larger reward differences. 

\paragraph{Robustness of DPO Training.}
Notably, none of the preference pairs degrades the performance of the SFT checkpoint.
This confirms that DPO training remains robust to different preference pairs. 
Even for suboptimal preference pairs, model performance does not regress below the baseline established by the SFT checkpoint, highlighting the stability of DPO.

\subsection{Analysis}

\paragraph{Extending Reward Positions.}
To further explore the impact of preference data construction, we extend our data construction to include additional reward points at \(\mu \pm 4\sigma\) and \(\mu \pm 3\sigma\). 
Experiments conducted on Llama-3-8B-Instruct reveal that sample points at \(\mu + 4\sigma\) and \(\mu + 3\sigma\) show no significant difference from selecting max-reward responses. 
Similarly, \(\mu - 4\sigma\) and \(\mu - 3\sigma\) exhibit no substantial difference from selecting min-reward responses. 
These findings suggest that expanding the reward range beyond \(\mu \pm 2\sigma\) does not provide additional benefits for preference optimization, reinforcing the sufficiency of our selected reward points.
The experimental results can be found in Appendix~\ref{appendix_extend}.

\paragraph{Scaling to 400 Samples Per Prompt.}
While our main experiments use 200 samples per prompt due to computational constraints, we also evaluate the scalability of our findings by conducting experiments with 400 samples per prompt. 
Based on the experiment with Llama-3-8B-Instruct as the SFT model, we observe that our conclusions remain consistent across both sample sizes.
More details on these results are provided in Appendix~\ref{appendix_400}.  

\paragraph{Training Dynamics and Loss Analysis.}
To better understand how different preference pairs influence DPO training, we record the training loss of six datasets, corresponding to the pairs \((max, min)\), \((max, \mu - 2\sigma)\), \((max, \mu - \sigma)\), \((max, \mu)\), \((max, \mu + \sigma)\), and \((max, \mu + 2\sigma)\). 
The loss curves, presented in Figure~\ref{loss}, reveal several important trends. 
First, larger reward margins facilitate training by enabling the model to converge more effectively. 
Models trained with larger reward gaps achieve lower loss values, which correlate with improved performance. 
By contrast, training loss for the pair \((max, \mu + 2\sigma)\) stagnates, indicating underfitting. 
We assume the reason is that it is difficult for the model to distinguish the chosen and rejected responses in this pair, leading to ineffective optimization.  
Interestingly, the preference dataset \((max, min)\) exhibits the lowest training loss. 
While this may suggest faster convergence, it also raises concerns about overfitting, as models trained on this dataset fail to perform as well as those trained with intermediate reward pairs. These findings highlight the trade-off between reward margins, optimization efficiency, and generalization performance. 
A more detailed empirical and theoretical analysis is provided in Appendix~\ref{overfitting}.

\section{Scaling Samples to Improve Alignment }
\label{main_exp}

\begin{figure*}[t]
\centering
\subfloat[Llama]{%
\begin{minipage}[t]{0.48\linewidth} 
    \centering
    \includegraphics[width=0.48\linewidth]{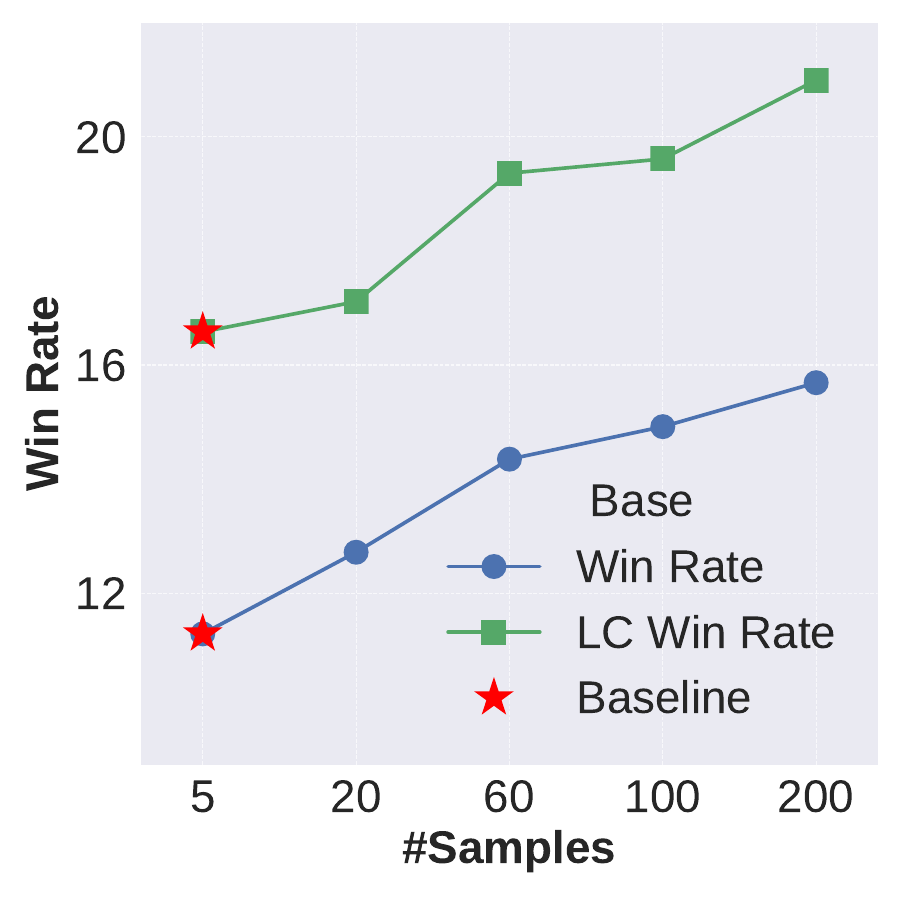}
    \hfill
    \includegraphics[width=0.48\linewidth]{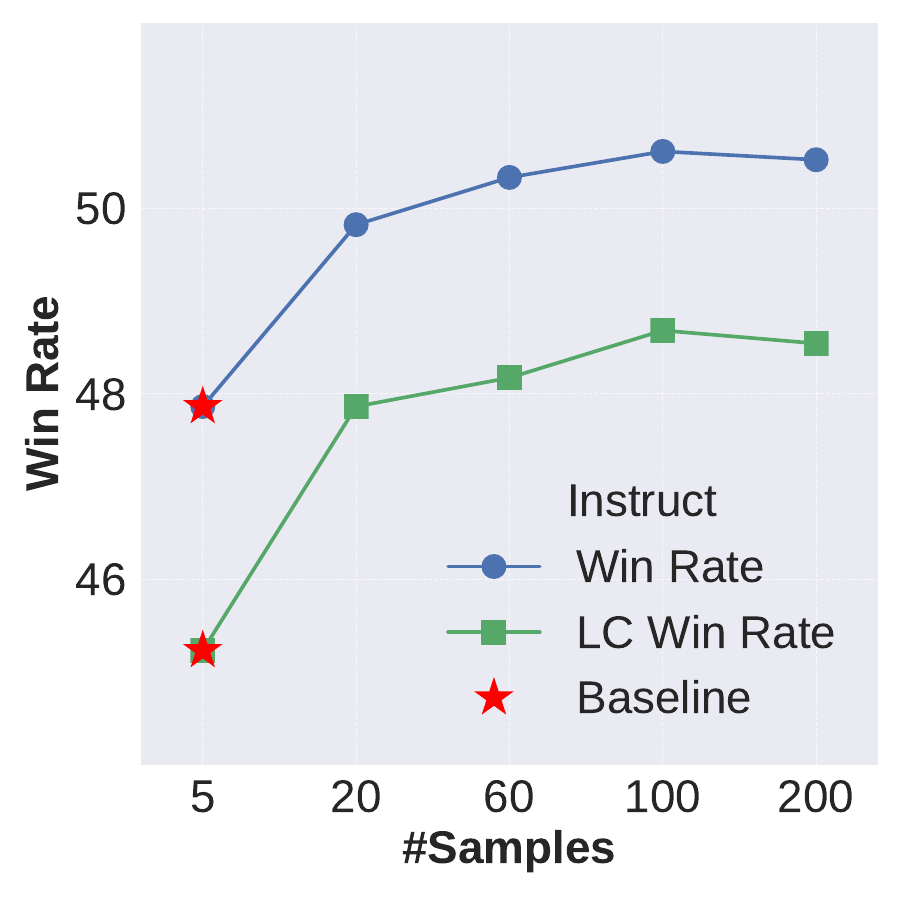}
\end{minipage}%
}
\hfill
\subfloat[Mistral]{%
\begin{minipage}[t]{0.48\linewidth} 
    \centering
    \includegraphics[width=0.48\linewidth]{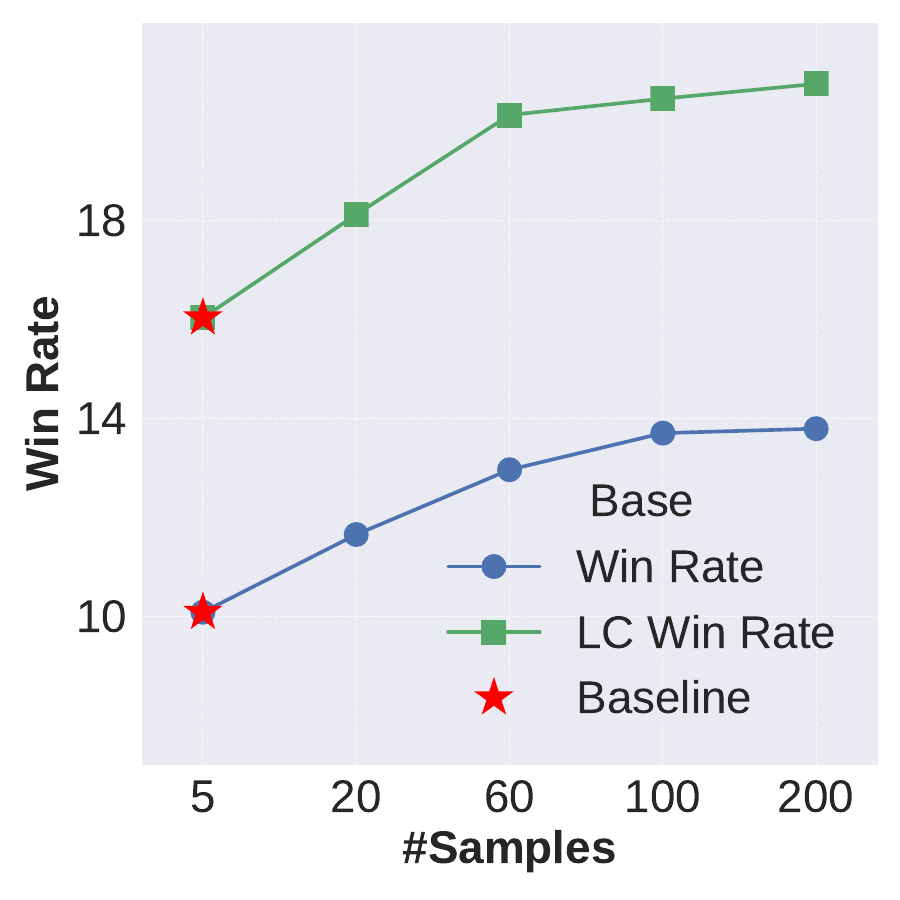}
    \hfill
    \includegraphics[width=0.48\linewidth]{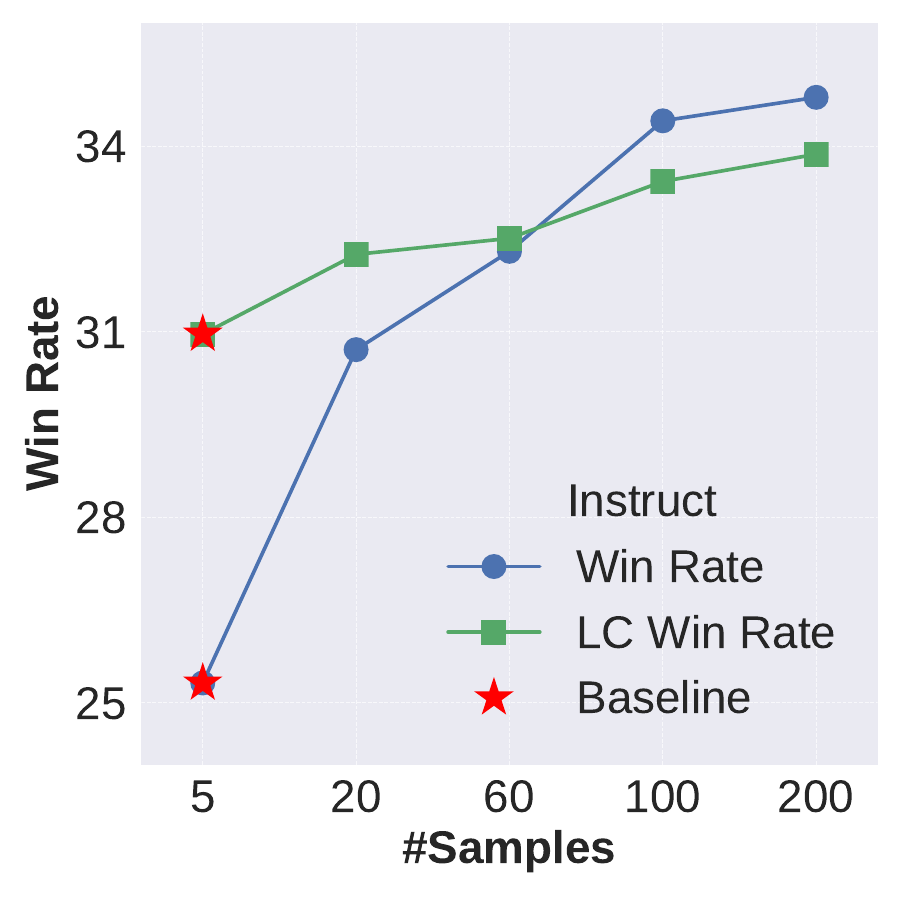}
\end{minipage}%
}
\caption{Alpaca evaluation results. The rejected response is selected as the one of the minimal rewards in 5 samples, while the chosen response is selected as the one of the maximal rewards in $n$ samples. We can improve the performance of models when increasing n within a range. x-axis is the number of samples ($n$), and y-axis is the performance.}
\label{fix_wr}
\end{figure*}

\begin{figure}[t]
\centering
\includegraphics[width=0.55\linewidth]{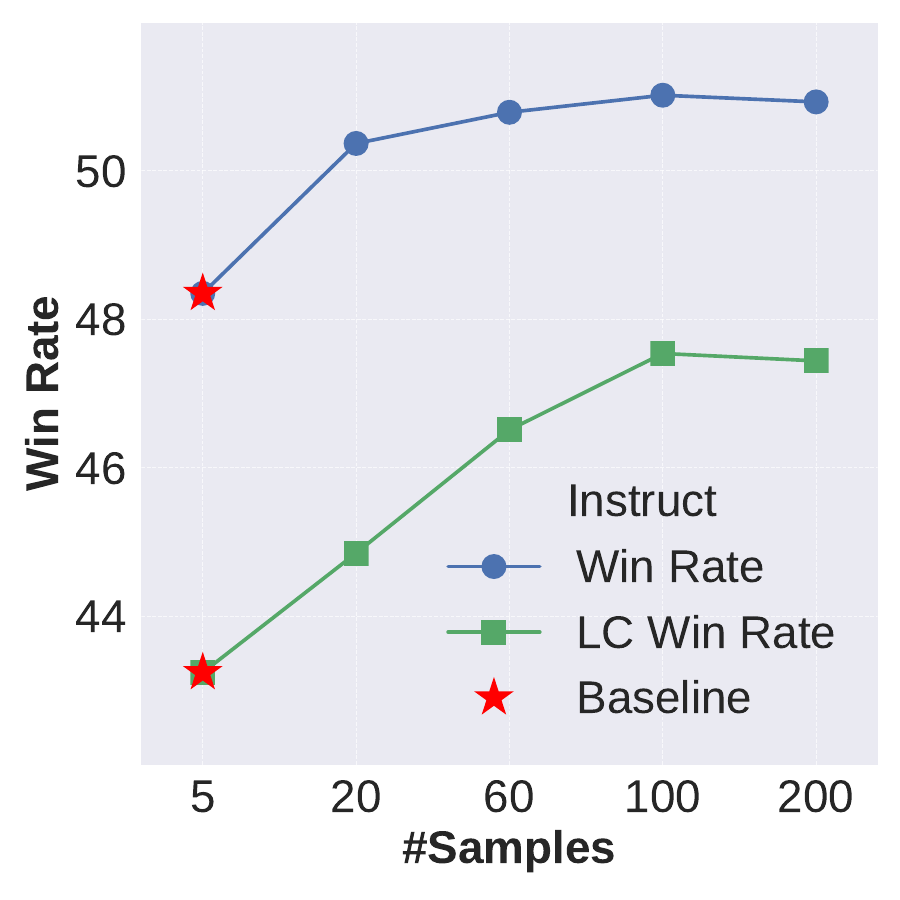}
\caption{Alpaca evaluation results. We demonstrate the effectiveness of preference data construction strategy on the Skywork reward model. x-axis is the number of samples ($n$), and y-axis is the performance.}
\label{llama_skywork}
\vspace{-1em}
\end{figure}

We established that selecting the rejected response at reward position \(\mu - 2\sigma\) is a key factor and model performance increases as the quality of the chosen response improves. 
Building on these insights, we propose a simple and effective preference pair construction strategy for DPO.
We further validate the effectiveness of this strategy across multiple reward models to ensure its robustness.  

\subsection{Data Construction Strategy}
Given a language model policy, a reward function, and \( k \) prompts \( \{x_i\}_{i=1}^k \), we sample \( n \) responses per prompt, denoted as \( \{y_{ij}\}_{j=1}^n \), from the policy model \( \pi_{\theta} \). 
Each response is scored using the reward function.
For the rejected response, we select the response with the lowest reward from 5 random samples. 
We find this approach to be an effective proxy for \(\mu - 2\sigma\) if the sample size is insufficient to approximate the true normal distribution of rewards. 
For the chosen response, we select the response with the highest reward among all \( n \) samples.  
This ensures that as \( n \) increases, the quality of the chosen responses improves naturally, leading to better preference optimization. 
An illustration of the data construction process is provided in Figure~\ref{pipe}. 
We further analyze how the quality of the chosen responses evolves with increasing sample size in Appendix~\ref{appendix_reward}. 

\subsection{Experimental Setup}
We evaluate our proposed preference data construction method by comparing it with the conventional approach, where the chosen response is selected as the one with the highest reward and the rejected response is the one with the lowest reward among five samples. 

For our method, we begin by sampling 5 responses per prompt. 
The response with the lowest reward is designated as the rejected response. 
As we progressively increase the number of sampled responses, we continue to select the chosen response as the one with the highest reward among all available candidates. 
This approach ensures that as the sample size grows, the quality of the chosen response improves, allowing us to examine the impact of a larger sample pool on model alignment. 
All experiments are conducted by following the implementation details outlined in Section~\ref{imp_detail}, unless specified otherwise.  

\subsection{Experimental Results and Analysis}

\paragraph{Scaling the Number of Samples.}  
The results of our proposed preference data construction strategy are presented in Figure~\ref{fix_wr}. 
For reference, the first point in each line represents the performance of the conventional approach. 
Since our method is only identical to the baseline when using five samples per prompt, performance differences emerge as \( n \) increases. 
We observe a steady improvement in performance across all models as we increase the number of samples from 5 to 200, even though the rate of improvement may diminish in some cases. 
The only exception occurs in Llama-3-8B-Instruct, where performance experiences a slight drop when increasing the number of samples from 100 to 200.


\paragraph{Comparison with Prior Work.}  
To further validate the effectiveness of our method, we compare it with the results of \citet{meng2024simpo} (first 2 rows) in Table~\ref{compare_literature}, which employs the conventional data construction strategy.
Our method can outperform baseline results of DPO in both benchmarks, AlpacaEval 2 and Arena-hard.
Furthermore, it can also surpass the baseline results of SimPO in terms of Alpaca win rate and Arena-hard win rate.  

\begin{table}[t]
\centering
\small
\begin{tabular}{lcccc}
\toprule
\textbf{Data (Method)} & \textbf{\#Sample} & \makecell{\textbf{AE}\\\textbf{LC}} & \makecell{\textbf{AE}\\\textbf{WR}} & \makecell{\textbf{AH}\\\textbf{WR}} \\
\midrule
Baseline\textsuperscript{*}(SimPO) & 5 & \textbf{53.7} & 47.5 & 36.5 \\
Baseline\textsuperscript{*}(DPO) & 5 & 48.2 &	47.5 & 35.2 \\
Baseline\textsuperscript{†}(SimPO) & 400 & 44.6 & 43.9 &  34.8 \\
Baseline\textsuperscript{†}(DPO) & 400 & 42.0 & 42.0 &  34.5 \\
\midrule
Ours (DPO) & 400 & 49.1 & \textbf{50.2} & \textbf{37.3} \\
\bottomrule
\end{tabular}
\caption{We compare our method with reported baseline scores from  ~\citet{meng2024simpo} on Llama-3-8B-Instruct. 
AE denotes alpaca evaluation. AH represents arena-hard evaluation~\cite{li2024crowdsourceddatahighqualitybenchmarks}.
LC denotes length controlled win rate, while WR denotes win rate.
* means the original results from ~\citet{meng2024simpo}.
† means our own implementation.}
\label{compare_literature}
\vspace{-1em}
\end{table}

\paragraph{Evaluation on Skywork Reward Model.}  
While our previous experiments used Armorm as the reward model, we also evaluate our preference data construction strategy using the Skywork reward model to ensure its general applicability. 
We adopt Llama-3-8B-Instruct as the SFT model and record the results of AlpacaEval 2 in Figure~\ref{llama_skywork}. 
We can see that model performance improves as the number of samples increases before reaching a plateau, which confirms that our strategy is robust across different reward models.  

\paragraph{Evaluation on Academic Benchmarks.}
To assess whether our preference data construction method negatively affects performance on established NLP benchmarks, we evaluate our trained model based on Llama-3-8B-Instruct on a set of widely used academic tasks, including ARC~\cite{clark2018thinksolvedquestionanswering}, HellaSwag~\cite{zellers-etal-2019-hellaswag}, TruthfulQA~\cite{lin-etal-2022-truthfulqa}, and GSM8K~\cite{cobbe2021trainingverifierssolvemath}. 
We use the Language Model Evaluation Harness~\cite{eval-harness} for evaluation. 
More details of our results are presented in Appendix~\ref{aca_bm}. 
We observe that our policy models do not show performance drops on academic benchmarks.

\section{Related Work}

\paragraph{Reinforcement Learning from Human Feedback.} 
RLHF is a dominant approach to align large language models with human preferences in the generation of natural language~\cite{NEURIPS2022_b1efde53, touvron2023llama2openfoundation}. 
The RLHF process generally involves three stages: initial supervised fine-tuning, reward modeling~\cite{lambert2024rewardbenchevaluatingrewardmodels}, and policy optimization. 
This approach has been extended to address various challenges such as reducing toxicity, improving safety and reasoning capabilities~\cite{qi2024safetyalignmentjusttokens, wu2023finegrained, dai2024safe, yu2024metamath}. 
However, RLHF has issues such as training instability and complexities due to the nature of reinforcement learning and multi-stage pipeline, potentially leading to biases and verbose model outputs.

DPO~\cite{rafailov2023direct} and its variants~\cite{meng2024simpo, pmlr-v235-ethayarajh24a, han2024fpogeneralizingpreferenceoptimization} were developed to overcome these limitations by directly integrating preferences into the policy model using pairwise preference data. 
This approach simplifies the policy optimization process by eliminating the need for a surrogate reward phase.
As an increasing number of powerful reward models become publicly available~\cite{jiang-etal-2023-llm, wang-etal-2024-interpretable, wang2024arithmetic, liu2024skyworkrewardbagtricksreward}, a popular practice~\cite{dong2023raft, liu2024statistical, meng2024simpo, ye-ng-2024-preference} has gained focus for models to enhance capability training through DPO, which employ reward models to select self-generated samples.

\paragraph{Synthetic Data for LLMs.} 
Human-curated data has consistently been a highly effective resource to enhance model performance in natural language processing~\cite{bai2022traininghelpfulharmlessassistant, kpf2023openassistant}.
Although human-curated data are typically of high quality, obtaining sufficient amounts can be prohibitively expensive. 
To address this challenge, the use of synthetic data has gained attention as a cost-effective alternative to human data~\cite{west-etal-2022-symbolic, hsieh-etal-2023-distilling, wang-etal-2023-self-instruct, dong2024selfboostinglargelanguagemodels, li2024selfalignment}.
This approach often involves the use of advanced LLMs to generate high-quality synthetic datasets~\cite{tajwar2024preferencefinetuningllmsleverage, dong2023raft, agarwal2024onpolicy, chen2025longpo}.
In particular, on-policy data has emerged as a highly effective and efficient approach, drawing considerable interest in recent work.
Previous work mainly aims to generate more high-quality data to improve the capabilities of LLMs.
In this paper, we focus on how to use self-generated data to construct optimal preference pairs for DPO.

\paragraph{Scaling Inference.}
Recently, many studies have explored how scaling inference (sample budgets) impacts the performance of large language models~\cite{wu2024scaling, brown2024largelanguagemonkeysscaling, snell2024scalingllmtesttimecompute, zhang2024scalingllminferenceoptimized}. 
They have shown that increasing the number of samples improves the solve rate with a simple best-of-N strategy~\cite{amini2024variationalbestofnalignment} and a powerful reward. 
Specifically, improvements in the performance of these models in mathematical problems have been achieved by repeatedly sampling potential solutions by manipulating temperatures. 
Research has also been conducted to investigate the scaling effects related to various inference algorithms, reward functions, and model sizes~\cite{chi2024thoughtsculptreasoningintermediaterevision, wu2024scaling}.
\section{Conclusion}
\label{conclusion}


We first point out the failure case of a conventional preference data construction strategy for DPO. 
To address this,  we then classify the samples into seven categories based on the underlying normal distribution of rewards per prompt and construct 21 preference datasets as a whole to systematically evaluate their impact on model performance.  
Our findings demonstrate that selecting the rejected response at reward position $\mu - 2\sigma$ is critical for effective optimization of DPO.
We finally propose a simple preference data construction strategy that can steadily improve the performance of trained models through DPO as the sample size increases.

\section*{Limitations}

Our method assumes the existence of a strong reward model, which may not always be readily available or easily trainable for all tasks. 
The quality of the reward model directly impacts the effectiveness of our approach, and inaccuracies in the reward model can lead to suboptimal performance or biased outcomes.
Another limitation is the computational cost associated with generating a large number of responses for each prompt to construct the preference dataset. 
For tasks with a large input space or high complexity, this can become resource-intensive and time-consuming.
We mainly focus on DPO in this paper and will explore other preference optimization methods in the future.

\section*{Acknowledgements}
This research/project is supported by the National Research Foundation, Singapore under its National Large Language Models Funding Initiative (AISG Award No: AISG-NMLP-2024-004). Any opinions, findings and conclusions or recommendations expressed in this material are those of the author(s) and do not reflect the views of National Research Foundation, Singapore. This research/project is supported by the Ministry of Education, Singapore, under its SUTD-SMU Joint Grant Call, if applicable. We also thank Professor Lu Wei for providing GPUs so that we can carry out preliminary experiments.

\bibliography{anthology,custom}

\appendix

\section{Appendix}

\subsection{Hyperparameters for Training}
\label{appendix_hyper}

Here, we report the details of our training. 
The learning rate and batch size are listed in Table~\ref{hyperparameters}. 
Our experiments are running on 8 H100 and 8 A100 GPUs.

\begin{table}[!ht]
\centering
\resizebox{0.48\textwidth}{!}{
\begin{tabular}{lcccccc}
\toprule
\multirow{2}{*}{\textbf{Hyperparameters}} & \multicolumn{2}{c}{\textbf{Llama-Base}} & \multicolumn{2}{c}{\textbf{Mistral-Base}} & \textbf{Llama-Inst} & \textbf{Mistral-Inst} \\
\cmidrule(lr){2-3} \cmidrule(lr){4-5} \cmidrule(lr){6-6} \cmidrule(lr){7-7}
& \textbf{SFT} & \textbf{DPO} & \textbf{SFT} & \textbf{DPO} & \textbf{DPO} & \textbf{DPO}\\
\midrule
Batch size       & 128  & 128   & 128  & 128  & 128  & 128\\
Epochs           & 1    & 1     & 1    & 1    & 1    & 1  \\
Learning rate    & 2e-5 & 5e-7  & 2e-5 & 3e-7 & 3e-7 & 3e-7\\
Beta             & -    & 0.01  & -    & 0.01 & 0.01 & 0.1\\
Warm-up ratio    & 0.1  & 0.01  & 0.1  & 0.01 & 0.01  & 0.01\\
\bottomrule
\end{tabular}
}
\caption{Hyperparameters for SFT and DPO training.}
\label{hyperparameters}
\end{table}

\subsection{Baseline Results}
\label{baseline}

For comparison, we report the results of SFT models and models trained with the conventional preference data construction strategy in Table~\ref{results_main_armorm}.
In base setting, the SFT models are trained on UltraChat, while it is the original instruct model in the instruct setting.
For the conventional preference data construction strategy, we sample five responses for each prompt.

\subsection{Win Rate Results}
\label{wr}

\begin{figure*}[!ht]
\centering

    \begin{minipage}[c]{\linewidth}
        \centering
        \includegraphics[width=0.48\linewidth]{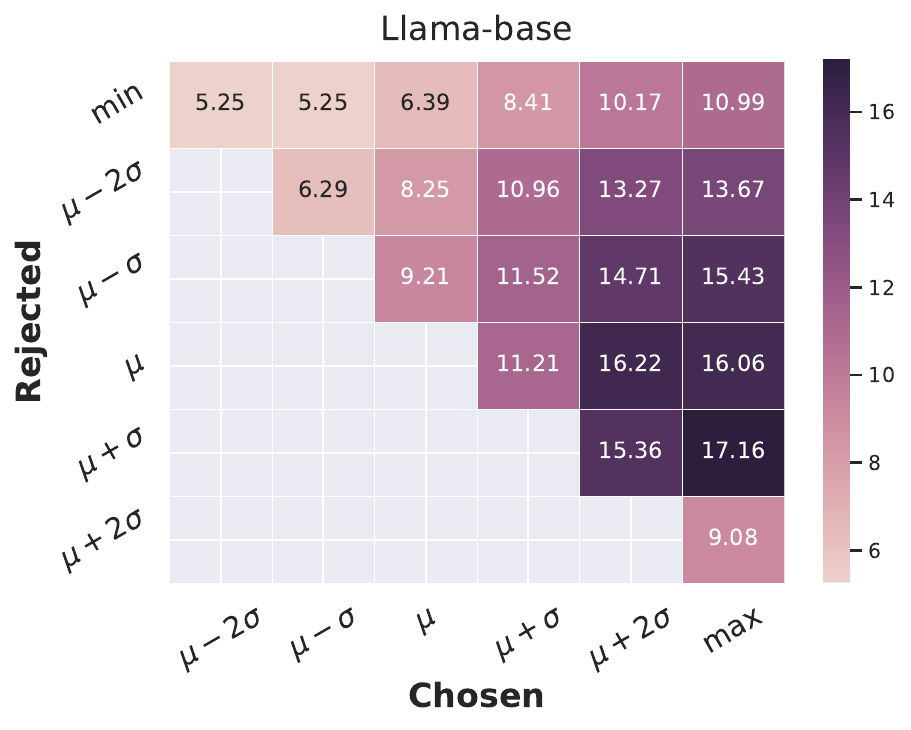}
        \hfill
        \includegraphics[width=0.48\linewidth]{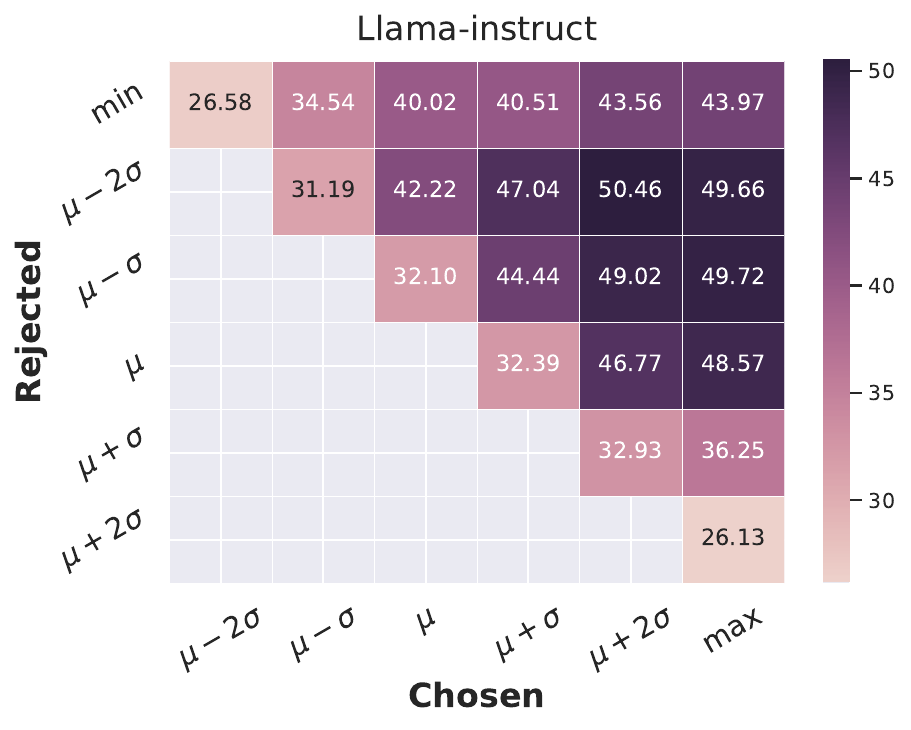}
        \\
        \includegraphics[width=0.48\linewidth]{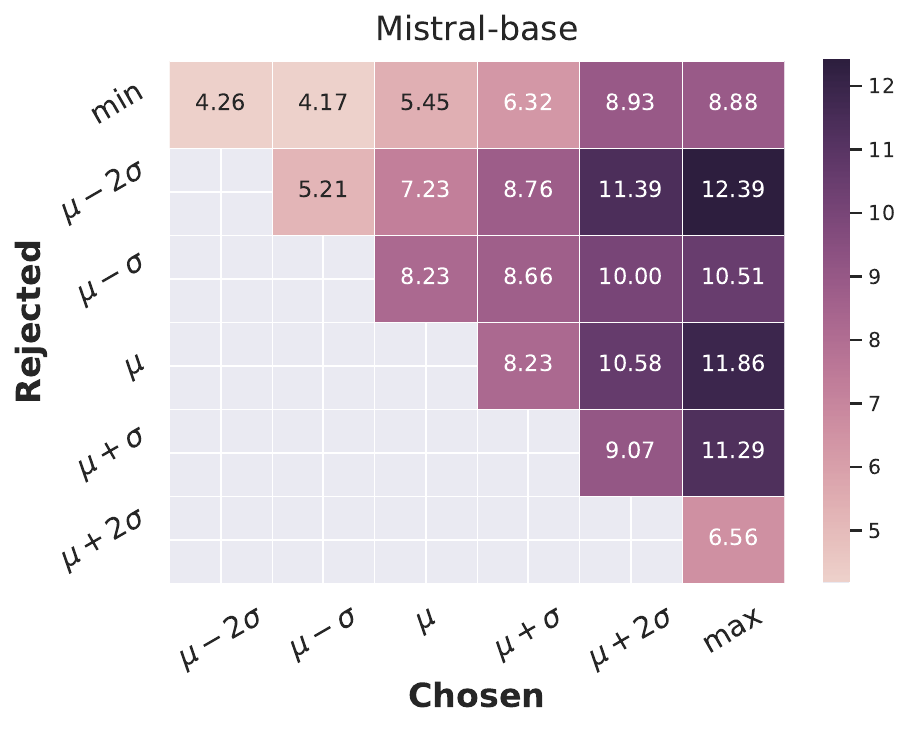}
        \hfill
        \includegraphics[width=0.48\linewidth]{figs/llama_instruct_wr.pdf}
    \end{minipage}

\caption{Win rate results of Alpaca evaluation.}
\label{main_fig_wr}
\end{figure*}

We report the win rate results of our policy models, which correspond to Section~\ref{data_cons} in Figure~\ref{main_fig_wr}. We find that our findings in Section~\ref{data_cons} also hold in most cases.

\subsection{Extend Reward Points}
\label{appendix_extend}

We have tried data construction based on the values $\left\{min, \mu \pm 2\sigma, \mu\pm\sigma, \mu, max\right\}$.
Here, we extend this set to further include the values $\left\{\mu \pm 4\sigma, \mu \pm 3\sigma\right\}$. 
We experiment on Llama-3-8B-Instruct. 
We find that sample points on $\left\{\mu + 4\sigma, \mu + 3\sigma\right\}$ have no significant differences from $\left\{max\right\}$ when used to train models through DPO.
In addition, sample points on $\left\{\mu - 4\sigma, \mu - 3\sigma\right\}$ have no significant differences from $\left\{min\right\}$.
We report some of our results in Table~\ref{wide_scale}.

\subsection{400 Samples per Prompt}
\label{appendix_400}
In this part, we experiment with 400 samples per prompt to explore whether our results of 200 samples can still hold.
As we have emphasized, we are given sufficient sample budgets.
The SFT model is Llama-3-8B-Instruct.
We report the results in Table~\ref{results_main_armorm_400}.
To save some computation and evaluation costs, we only train and evaluate 11 models.
Our conclusions of constructed datasets are consistent for both scenarios, 200 samples and 400 samples per prompt.

\begin{table}[t]
\centering
\resizebox{0.35\textwidth}{!}{
\begin{tabular}{l l c c }
\toprule
\multirow{2}{*}{\textbf{Chosen}} & \multirow{2}{*}{\textbf{Rejected}} & \multicolumn{2}{c}{\textbf{Llama-Inst}} \\
\cmidrule(lr){3-4} & & \textbf{LC} & \textbf{WR} \\
\midrule         
$max$          & $min$                              & 43.11      & 43.97         \\
\midrule
$max$          & $\mu-4\sigma$                      & 43.61      & 44.30         \\
$max$          & $\mu-3\sigma$                      & 44.10      & 43.78         \\
\midrule
$max$          & $\mu-2\sigma$                      & 48.12      & 49.66                \\
\midrule
$\mu+4\sigma$  & $\mu-2\sigma$                      & 48.11      & 49.46         \\
$\mu+3\sigma$  & $\mu-2\sigma$                      & 48.26      & 49.20         \\
\bottomrule
\end{tabular}
}
\caption{Evaluation results of extended data points.}
\label{wide_scale}
\vspace{-1em}
\end{table}

\begin{table}[!t]
\centering
\resizebox{0.35\textwidth}{!}{
\begin{tabular}{l l c c}
\toprule
\multirow{2}{*}{\textbf{Chosen}} & \multirow{2}{*}{\textbf{Rejected}} & \multicolumn{2}{c}{\textbf{Llama-Inst}} \\
\cmidrule(lr){3-4} & & \textbf{LC} & \textbf{WR} \\
\midrule         
${none}^{*}$    & ${none}^{*}$                               & 16.58      & 11.29         \\
$max\_of\_5$    & $min\_of\_5$                               & 45.23      & 47.86         \\
\midrule
$max$           & $min$                                      & 42.01      & 42.04         \\
\rowcolor[gray]{0.9}
$max$           & $\mu-2\sigma$                              & 49.09      & 50.23         \\
$max$           & $\mu-\sigma$                               & 48.27      & 49.76         \\
$max$           & $\mu$                                      & 48.21      & 48.76         \\
$max$           & $\mu+\sigma$                               & 42.32      & 38.35         \\
$max$           & $\mu+2\sigma$                              & 28.42      & 27.81         \\
$\mu+2\sigma$   & $min$                                      & 39.35      & 40.08         \\
\rowcolor[gray]{0.9}
$\mu+2\sigma$   & $\mu-2\sigma$                              & 50.71      & 51.37         \\
$\mu+2\sigma$   & $\mu-\sigma$                               & 48.41      & 49.56         \\ 
$\mu+2\sigma$   & $\mu$                                      & 46.79      & 45.39          \\
$\mu+2\sigma$   & $\mu+\sigma$                               & 35.15      & 32.57         \\

\bottomrule
\end{tabular}
}
\caption{400 Samples per prompt data construction results on Alpaca evaluation.}
\label{results_main_armorm_400}
\vspace{-1em}
\end{table}

\begin{table*}[t]
\centering
\resizebox{0.9\textwidth}{!}{
\begin{tabular}{l l c c c c c c c c c c}
\toprule
\multirow{2}{*}{\textbf{Chosen}} & \multirow{2}{*}{\textbf{Rejected}} & \multicolumn{2}{c}{\textbf{Llama-Base}} & \multicolumn{2}{c}{\textbf{Mistral-Base}} & \multicolumn{2}{c}{\textbf{Llama-Inst}} & \multicolumn{2}{c}{\textbf{Mistral-Inst}} \\
\cmidrule(lr){3-4} \cmidrule(lr){5-6} \cmidrule(lr){7-8} \cmidrule(lr){9-10}
& & \textbf{LC} & \textbf{WR} & \textbf{LC} & \textbf{WR} & \textbf{LC} & \textbf{WR} & \textbf{LC} & \textbf{WR} \\
\midrule         
${none}^{*}$    & ${none}^{*}$    & \phantom{0}7.50   & \phantom{0}4.34  & \phantom{0}6.90   & \phantom{0}4.32           & 23.75       & 24.23       & 17.63       & 14.68       \\
$max\_of\_2$    & $min\_of\_2$    & 10.69      & \phantom{0}6.83       &          15.12   &       \phantom{0}8.36      &    31.86    &   32.61     &         21.97    &   17.55     \\
$max\_of\_5$    & $min\_of\_5$    & 16.58      & 11.29      & 16.03       & 10.08       & 45.23       & 47.86       & 30.96       & 25.32       \\

\bottomrule
\end{tabular}
}
\caption{We report the results of baselines, SFT models, and DPO models trained with the conventional data construction strategy. The method with $*$ denotes the results of SFT model in the base setting and instruct model in the instruct setting.}
\label{results_main_armorm}
\vspace{-1em}
\end{table*}
\begin{figure}[!t]
\centering
    \begin{minipage}[c]{\linewidth}
        \centering
        \includegraphics[width=0.9\linewidth]{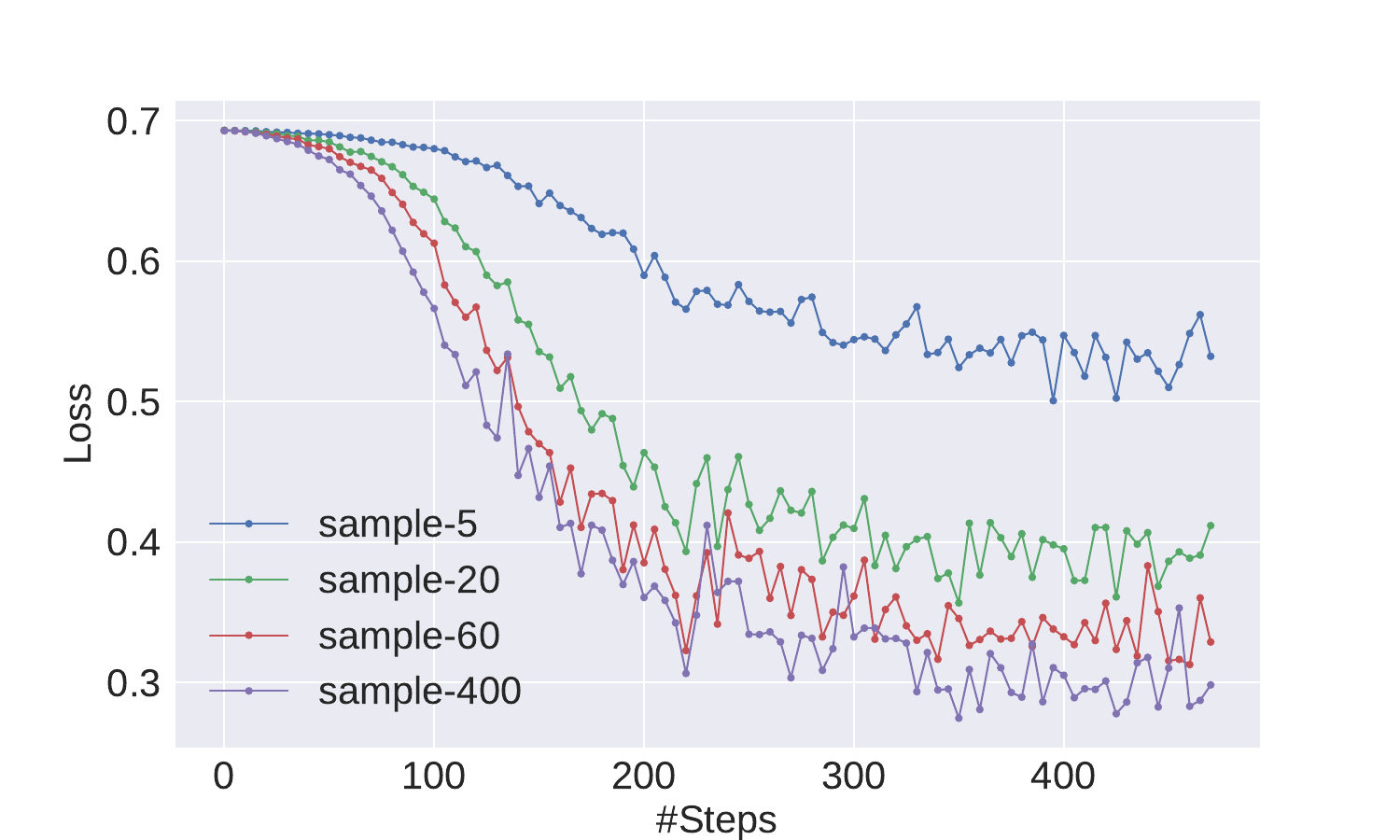}
    \end{minipage}
\caption{Loss of Llama-3-8B-Instruct when training with the data by selecting the response of maximal reward as the chosen response and selecting the response of minimal reward as the rejected response among $n$ responses.}
\label{ana_loss}
\end{figure}
\begin{figure}[t]
\centering
    
    \includegraphics[width=0.7\linewidth]{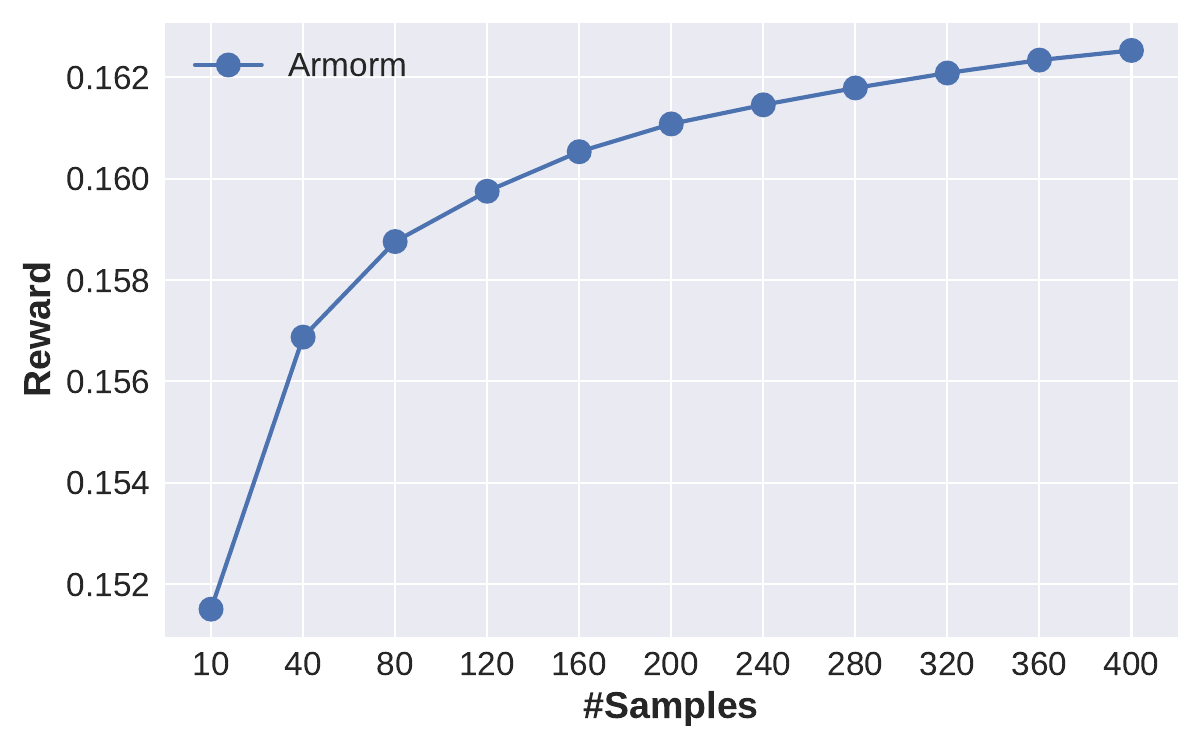}
    \hspace{0.05\linewidth}
    \\
    
    \includegraphics[width=0.7\linewidth]{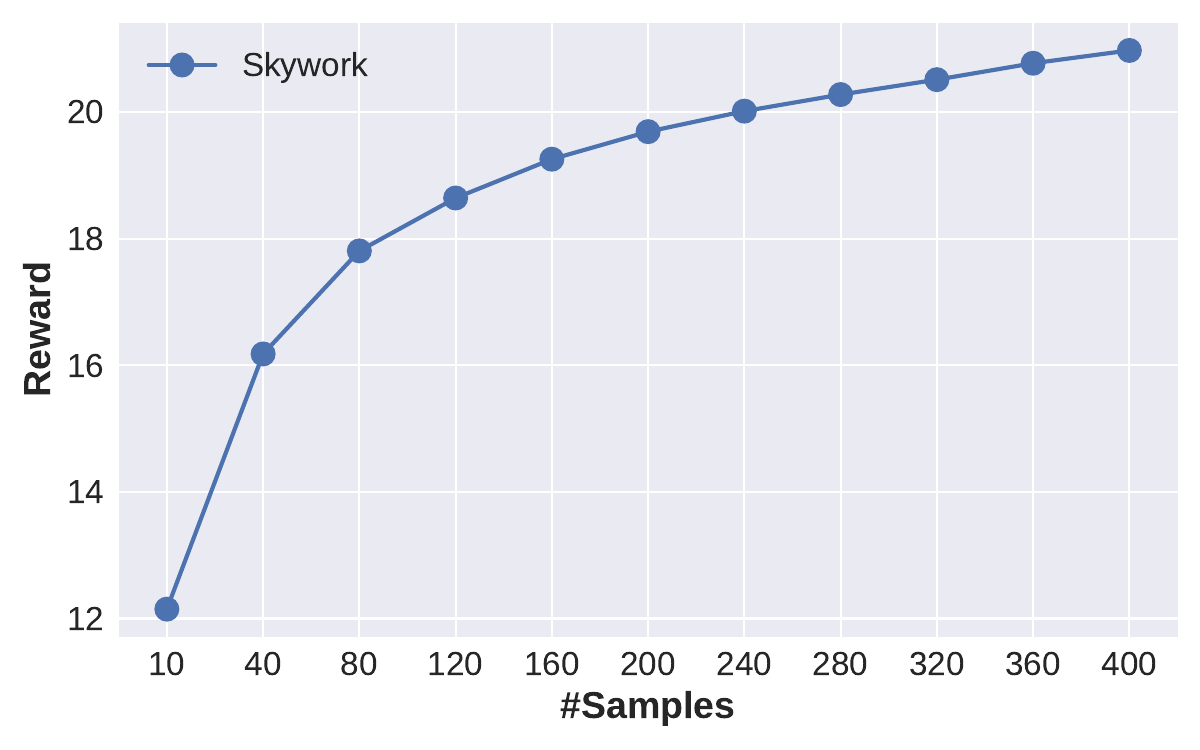}
    \caption{The average reward of 3 top ranking responses. }
    \label{reward} 
    \vspace{-1em}
\end{figure}

\subsection{Overfitting}
\label{overfitting}

We find that the data construction strategy, which selects the generation of maximal reward as the chosen response and selects the one with minimal reward as the rejected response among $n$ responses, may lead to overfitting of policy models when $n$ reaches a point.
\paragraph{Empirical Results.} As shown in Figure~\ref{ana_loss}, although the training loss can reach a lower bound when we increase $n$ from 5 to 400, the length-controlled win rate does not improve accordingly, 45.23, 46.88, 42.98, 42.01 for 5, 20, 60, 400 samples, respectively.
 
\paragraph{Theoretical Support.}

The reward model produces scores that are approximately normally distributed:
\[
r(y) \sim \mathcal{N}(\mu, \sigma^2).
\]
When evaluating a large number \( n \), the extreme values—namely, the maximum \( \max_{j=1}^{n} r(y_j) \) and the minimum \( \min_{j=1}^{n} r(y_j) \)—tend to be statistical outliers rather than typical examples from the distribution. According to extreme value theory, for large \( n \), the expected maximum and minimum rewards are approximately given by
\[
\mathbb{E}\left[ \max_{j=1}^{n} r(y_j) \right] \approx \mu + \sigma \sqrt{2 \log n},
\]
\[
\mathbb{E}\left[ \min_{j=1}^{n} r(y_j) \right] \approx \mu - \sigma \sqrt{2 \log n}.
\]

In the DPO framework, the log-likelihood loss is sensitive to the differences in reward scores. Specifically, consider the term
\[
\log \sigma \bigl(r(y_w) - r(y_l)\bigr),
\]
where \( r(y_w) \) and \( r(y_l) \) represent the rewards for the winning and losing samples, respectively. Using the approximations above, the difference between an extreme high and an extreme low reward scales roughly as
\[
r(y_w) - r(y_l) \approx 2\sigma \sqrt{2 \log n},
\]
so that
\[
\log \sigma \bigl(r(y_w) - r(y_l)\bigr) \approx \log \sigma \bigl(2\sigma \sqrt{2 \log n}\bigr).
\]

As \( n \) increases, this term becomes saturated, which diminishes the effective learning signal. In other words, the presence of extreme outliers can lead the optimization process to overfit to these statistical artifacts rather than capturing improvements that generalize well.

\subsection{Reward with More Samples}
\label{appendix_reward}

To support our view that the quality of top-ranking responses is getting better, we record the average reward score of the three highest-ranking responses for 1,000 prompts in Ultrafeedback. 
As shown in Figure~\ref{reward}, the reward scores of both Armorm and Skywork reward models confirm our hypothesis that the quality of top-ranking responses improves as the number of samples increases.

\subsection{Evaluation on Academic Benchmarks}
\label{aca_bm}


\begin{table}[t]
\centering  
\resizebox{0.48\textwidth}{!}{%

\begin{tabular}{lccccc}
\toprule
\textbf{Tasks} & \textbf{ARC\_C(5)} & \textbf{HS(10)} & \textbf{TQA(0)} & \textbf{GSM(5)} \\
\midrule
Llama-Inst   & 57.25 &  58.71 & 35.99 & 75.06 \\
\midrule
Ours &   &  &  &  \\
5       & 61.43  & 59.19 & 40.64 & 76.88 \\
20      & 61.52  & 58.90 & 39.78 & 75.15 \\
60     & 61.52  & 58.79 & 39.66 & 76.19 \\
100     & 61.26  & 59.02 & 39.41 & 76.19 \\
200     & 61.43  & 58.84 & 39.66 & 77.26 \\
\bottomrule
\end{tabular}%
}
\caption{Performance of trained models with 5, 20, 60, 100, 200 samples per prompt on academic benchmarks. We observe no performance drops. HS denotes HellaSwag, while TQA denotes TruthfulQA.}
\label{task_performance}
\vspace{-1em}
\end{table}

In Table~\ref{task_performance}, we evaluate our trained model based on Llama-3-8B-Instruct on a set of widely used academic tasks, including ARC~\cite{clark2018thinksolvedquestionanswering}, HellaSwag~\cite{zellers-etal-2019-hellaswag}, TruthfulQA~\cite{lin-etal-2022-truthfulqa}, and GSM8K~\cite{cobbe2021trainingverifierssolvemath}.

\end{document}